\documentclass{article}

\PassOptionsToPackage{numbers,compress}{natbib}
\usepackage[main, final]{neurips_2026}

\usepackage[utf8]{inputenc} %
\usepackage[T1]{fontenc}    %
\usepackage[colorlinks,
            linkcolor=red,
            anchorcolor=blue,
            citecolor=blue, 
            pagebackref=true
           ]{hyperref}
\usepackage{url}            %
\usepackage{booktabs}       %
\usepackage{amsfonts}       %
\usepackage{nicefrac}       %
\usepackage{microtype}      %
\usepackage[dvipsnames]{xcolor} %

\usepackage{amsmath,amssymb,amsthm,bm,mathtools}

\usepackage{makecell}

\usepackage{caption}
\usepackage{algorithm}
\usepackage{algorithmic}
\usepackage{xcolor}
\usepackage{mmll}
\usepackage{wrapfig}
\usepackage{enumitem}

\newcommand{\E}{\mathbb{E}}

\newcommand{\bthetaold}{\bm{\theta}_{\mathrm{old}}}
\newcommand{\old}{\mathrm{old}}

\allowdisplaybreaks
\title{Ordered Policy Optimization}

\author{%
  Dongkyu Derek Cho\\
  Department of Statistical Science\\
  Duke University\\
  \texttt{dc430@duke.edu} \\
  \And 
  Pan Xu \\
  Department of Biostatistics \& Bioinformatics \\
  Department of Computer Science \\
  Department of Electrical \& Computer Engineering \\
  Duke University \\
  \texttt{pan.xu@duke.edu}
}

\begin{document}

\maketitle

\begin{abstract}
Proximal Policy Optimization (PPO) stabilizes policy updates by clipping the importance ratio between the new policy and the data-collecting policy. The standard rule uses one fixed clipping range for all samples. While this choice offers simplicity and effectiveness, it fails to distinguish statistically harmful extreme ratio values from large, yet tolerable changes. We study how to adapt the clipping radius to the reliability of the induced importance ratio distribution. Our case study demonstrates that reliable updates require controlling these extreme ratios, rather than merely verifying that individual ratios fall within a fixed interval. Motivated by this observation, we propose Ordered Policy Optimization (\textsc{OPO}). In each minibatch, \textsc{OPO} sorts the importance ratios, selects the extreme ratios that potentially have the largest effect on the update, and assigns them adaptive sample-wise radii instead of using the fixed clipping radius. These radii are derived from a prescribed tail profile, which specifies desired target values for the selected extreme ratios. The resulting objective remains first-order and requires only minibatch sorting and a modified policy loss. Experiments conducted on MuJoCo and Atari benchmarks demonstrate that this simple modification achieves strong performance relative to established baselines. Code is available at \url{https://github.com/CrawlingKiming/ordered_ppo}. 
\end{abstract}

\section{Introduction}
Deep reinforcement learning (RL) has achieved strong empirical performance in domains such as games, robotic control, and large-scale model fine-tuning \citep{sutton1999policy, schulman2015high, mnih2015human, silver2016mastering, schulman2017proximal}. Among these approaches, policy-gradient methods optimize a parametrized policy
through gradient-based updates of the expected return \citep{williams1992simple,sutton1999policy,konda1999actor,kakade2002approximately,schulman2017proximal,xu2019improved,xu2020sample,wu2020finite}. Their practical performance, however, depends critically on the stability of the policy update.

A fundamental challenge arises because policy updates typically evaluate a candidate policy using data collected under a previous policy \citep{SchulmanTRPO2015, schulman2017proximal}. This necessitates an importance-weighted surrogate objective, whose behavior is governed by the likelihood ratio between the candidate and behavior policies. Trust Region Policy Optimization (TRPO) addresses this issue by maximizing the surrogate objective under an explicit constraint on policy discrepancy \citep{SchulmanTRPO2015}. While theoretically robust, this constrained optimization can be computationally cumbersome for large-scale applications.

Proximal Policy Optimization (PPO) replaces TRPO's constrained optimization with a clipped first-order surrogate and has become a standard practical method \citep{schulman2017proximal,engstrom2020implementation,ouyang2022training,xie2024simple}. Its clipping rule acts pointwise on each sampled likelihood ratio using a fixed interval around one. Although effective, this rule does not directly control the tail behavior of the ratio distribution, which governs the reliability of the importance-weighted surrogate \citep{wang2020truly,vehtari2017practical,yao2018yes,vehtari2024pareto}.

To bridge this gap, we show that instability in updates based on the importance-weighted surrogate is driven by extreme likelihood ratios, rather than by the typical ratios found in the bulk of the distribution \citep{vehtari2024pareto}. Our motivating analysis shows that a sampled ratio falling within a fixed clipping band does not   
\begin{figure}[t]
\centering
\includegraphics[width=0.75\linewidth, trim={1.5cm 1.5cm 0.5cm 1.5cm},
        clip]{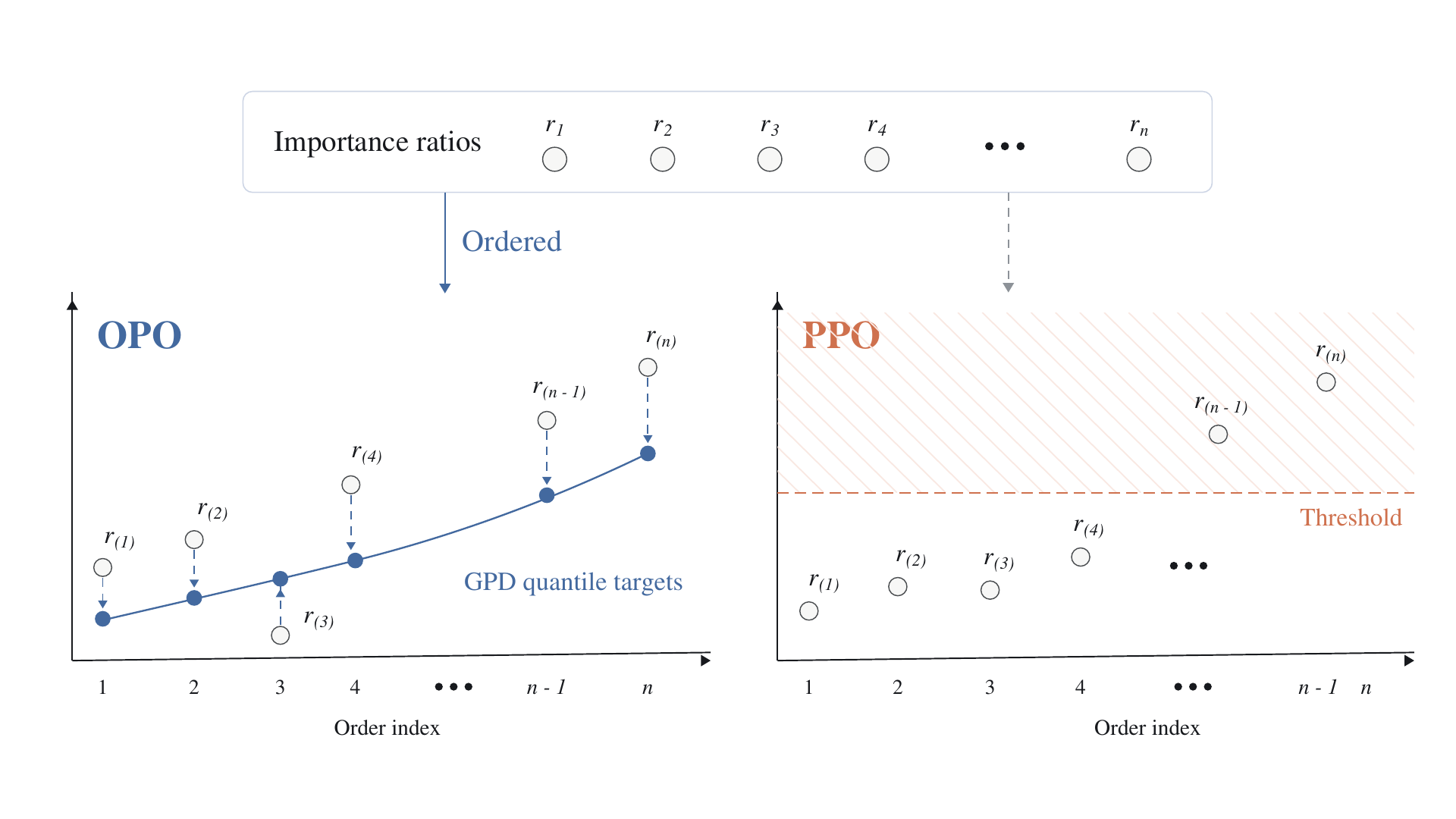}
\caption{
Overview of \textsc{OPO}. A minibatch of importance ratios is ordered, and the empirical tails are assigned teacher targets induced by a prescribed generalized Pareto tail profile. This yields adaptive sample-wise proximal targets, thereby replacing scalar clipping with order-based tail control. In contrast, PPO imposes the same scalar clipping boundary across ratios. 
}
\label{fig:opo_overview}
\end{figure}
necessarily imply meaningful control of policy discrepancy. Moreover, the tail parameter of the ratio distribution governs the moment regime of the policy-update estimator. This insight suggests that the relevant object for stabilization is not a local pointwise clipping event, but the ordered tail of the empirical ratio distribution. Based on this observation, we propose \emph{Ordered Policy Optimization} (OPO). OPO constructs teacher targets for the upper and lower empirical tails of the ratio distribution and translates these targets into adaptive sample-wise proximal radii. The resulting objective remains first-order and computationally simple, requiring only minibatch sorting and a modified proximal penalty. Thus, OPO preserves the computational efficiency of PPO while replacing fixed pointwise clipping with order-statistic-based tail control.

\textbf{Contributions:}
Our main contributions are summarized as follows.
\begin{itemize}
    \item We identify a statistical limitation of pointwise ratio clipping in proximal policy optimization. Through a motivating analysis, we show that the tail behavior of the likelihood ratio distribution governs the estimator regime of the importance-weighted policy surrogate.
    \item Motivated by this insight, we propose an order-statistic-based proximal objective. This objective regulates the empirical upper and lower tails of the ratio distribution by inducing adaptive, sample-wise radii derived from prescribed tail targets, yielding a method that is straightforward to implement.
    \item Empirically, we demonstrate that this simple modification yields strong performance on standard deep RL benchmarks, including MuJoCo and Atari.
\end{itemize}

\subsection{Additional Related Work}

\paragraph{Policy optimization.}
Policy-gradient methods optimize stochastic policies through gradient updates. Trust-region methods stabilize these updates by constraining the discrepancy between successive policies
\citep{sutton1999policy,kakade2002approximately,SchulmanTRPO2015,wang2019trust,wang2020truly}. PPO replaces this constrained optimization with a clipped first-order surrogate, making it one of the dominant practical methods in deep RL
\citep{schulman2017proximal,engstrom2020implementation,ouyang2022training,xie2024simple}. Several subsequent works extend PPO by adapting the clipping range or by replacing the clipped objective with alternative proximal surrogates \citep{chen2018adaptive, wang2019trust, wang2020truly, xie2024simple}. These methods are designed to improve the stability of policy updates while preserving the computational advantages of first-order optimization \citep{schulman2017proximal}. Recent studies have also considered simple proximal objectives that regulate likelihood ratios more directly \citep{chen2018adaptive, wang2019trust, wang2020truly, xie2024simple}. Our method differs from this line of work in its underlying motivation. Specifically, it is motivated by the observation that the statistical regime of the importance-weighted surrogate is governed by the tail behavior of the ratio distribution. Accordingly, the proposed adaptation is induced by order statistics, rather than by a pointwise clipping rule alone.

\paragraph{Importance weighting and extreme-value modeling.}
The instability of importance-weighted estimators has long been recognized in the statistical literature \citep{pickands1975statistical, vehtari2017practical, yao2018yes, vehtari2024pareto}. Extreme-value theory provides an asymptotic description of upper-tail behavior through threshold exceedances, which are approximated by a generalized Pareto law under suitable regularity conditions \citep{pickands1975statistical}. Pareto-smoothed importance sampling uses this perspective to diagnose and stabilize importance-weighted estimators, with the fitted tail parameter serving as a concise summary of estimator reliability. In that line of work, the role of extreme-value modeling is primarily diagnostic or post hoc \citep{vehtari2017practical, yao2018yes, vehtari2024pareto,cho2025fast}. Our use of this perspective is different. We do not fit a generalized Pareto model within each policy update, nor do we use tail estimates solely for diagnosis. Instead, we prescribe a tail profile, use it to construct order-statistic targets, and thereby obtain a simple proximal objective for policy optimization.

\section{Preliminaries}
\subsection{Reinforcement Learning and Trust-Region Policy Optimization}
We consider a discounted Markov decision process (MDP) \( \mathcal{M}=(\mathcal{S},\mathcal{A},P,\mathcal{R},\rho_0,\gamma) \), where \(\mathcal{S}\) and \(\mathcal{A}\) denote the state and action spaces, \(P(\bs' \mid \bs,\ba)\) is the transition kernel, \(\mathcal{R}(\bs,\ba)\) is the reward function, \(\rho_0\) is the initial-state distribution, and \(\gamma\in(0,1)\) is the discount factor. A stochastic policy \(\pi(\ba\mid\bs)\) induces a trajectory \(\btau=(\bs_0,\ba_0,r_0,\bs_1,\ldots)\). The learning objective is to maximize the expected discounted return $\eta(\pi) = \E \left\{ \sum_{t=0}^{\infty}\gamma^t \mathcal{R}(\bs_t,\ba_t) \right\}.$ %

When policy improvement is carried out using trajectories generated by a previous policy, value functions play a central role. Given a policy \(\pi\), we write \(V^\pi(\bs)=\E\{\sum_{t=0}^{\infty}\gamma^t \mathcal{R}(\bs_t,\ba_t)\mid \bs_0=\bs,\pi\}\) for the state-value function and \(Q^\pi(\bs,\ba)=\E\{\sum_{t=0}^{\infty}\gamma^t \mathcal{R}(\bs_t,\ba_t)\mid \bs_0=\bs,\ba_0=\ba,\pi\}\) for the action-value or Q function, with the advantage function defined as \(A^\pi(\bs,\ba)=Q^\pi(\bs,\ba)-V^\pi(\bs)\).

Value-based methods such as DQN improve sample efficiency by storing past transitions in a replay buffer and reusing them across updates \citep{mnih2015human}. Classical policy-gradient methods are instead on-policy; after a policy update, reusing trajectories from the previous policy requires correcting the behavior--target policy mismatch. The performance difference lemma \citep{kakade2002approximately} makes this policy-improvement problem explicit. For a reference policy \(\pi\) and a candidate policy \(\tilde{\pi}\), it gives $\eta(\tilde{\pi})-\eta(\pi)
=
\frac{1}{1-\gamma}
\E_{\bs\sim\rho_{\tilde{\pi}},\,\ba\sim\tilde{\pi}(\cdot\mid \bs)}
\!\{
A^\pi(\bs,\ba)\}$
where $\rho_{\tilde{\pi}}$ denotes the normalized discounted state visitation distribution under \(\tilde \pi\). Thus, relative to the reference policy \(\pi\), improvement is governed by whether \(\tilde{\pi}\) places more probability on actions with positive advantage \(A^\pi\). 

However, this expectation is taken under \(\rho_{\tilde{\pi}}\), which is unavailable before executing \(\tilde{\pi}\). A local surrogate therefore replaces \(\rho_{\tilde{\pi}}\) by \(\rho_\pi\), while the change in the conditional action distribution is accounted for by the likelihood ratio, yielding an objective that can be estimated from trajectories generated by \(\pi\). This surrogate is reliable only when the candidate policy remains close to the behavior policy. If \(\tilde{\pi}\) assigns substantial mass to actions that were unlikely under \(\pi\), the likelihood ratio can become large, causing a small number of samples to dominate the gradient estimate. Trust Region Policy Optimization (TRPO) \citep{SchulmanTRPO2015} implements this principle by maximizing an importance-weighted surrogate subject to an average KL-divergence constraint between the old and new policies: 
\begin{equation}
\max_{\btheta}\;
\E_{(\bs_t,\ba_t)}
\!\left\{
\frac{\pi_{\btheta}(\ba_t\mid \bs_t)}
     {\pi_{\bthetaold}(\ba_t\mid \bs_t)}
\hat{A}_t
\right\}
\quad
\text{s.t.}\quad
\E_{\bs_t\sim \pi_{\bthetaold}}
\!\left\{
D_{\mathrm{KL}}\!\bigl(
\pi_{\bthetaold}(\cdot\mid\bs_t)\,\|\,\pi_{\btheta}(\cdot\mid\bs_t)
\bigr)
\right\}
\le \delta,
\label{eq:trpo_obj}
\end{equation}
where \(\hat A_t\) estimates \(A_t:=A^{\pi_{\bthetaold}}(s_t,a_t)\). Directly solving \eqref{eq:trpo_obj} is difficult as it requires optimizing a high-dimensional policy subject to an empirical KL constraint. TRPO addresses this difficulty by linearizing the surrogate objective and quadratically approximating the KL constraint, which yields an approximate natural-gradient update. This approximation, however, still requires conjugate-gradient iterations, curvature estimation, damping, and a line search to satisfy the empirical KL bound \citep{SchulmanTRPO2015,schulman2017proximal}.

To alleviate these practical concerns, Proximal Policy Optimization (PPO) \citep{schulman2017proximal} replaces the hard trust-region constraint by a simpler first-order surrogate based on clipping the
sampled likelihood ratio \(
r_t(\btheta)
=
\pi_{\btheta}(\ba_t\mid\bs_t)/
\pi_{\bthetaold}(\ba_t\mid\bs_t)
\).
Its clipped objective is
\begin{align}
L^{\mathrm{clip}}(\btheta)
&=
\E_{(\bs_t,\ba_t)}
\!\big[
\min\!\bigl\lbrace
r_t(\btheta)\hat{A}_t,\,
\operatorname{clip}~\!\bigl(r_t(\btheta),1-\kappa,1+\kappa\bigr)\hat{A}_t
\bigr\rbrace
\big].
\label{eq:ppo_clip_obj}
\end{align}
This computational simplification delegates policy-update control entirely to the sampled ratio $r_t(\btheta)$ 
and the predefined scalar interval $[1-\kappa, 1+\kappa]$. However, a growing body of empirical and theoretical work indicates that PPO may deviate from its intended trust-region behavior, that implementation details can substantially influence performance, and that relying solely on ratio clipping may be insufficient to prevent ratio-induced instability \citep{wang2020truly, xie2024simple}. These observations lead to the following fundamental structural question:
\begin{center}
  What statistical properties of the likelihood ratio distribution are sufficient to ensure that the importance-weighted surrogate remains statistically reliable?
\end{center}

\begin{wrapfigure}[13]{r}{0.33\textwidth}
    \centering
    \vspace{-0.1in}
    \includegraphics[
        width=\linewidth,
        trim={0.4cm 1.2cm 0.4cm 2cm},
        clip
    ]{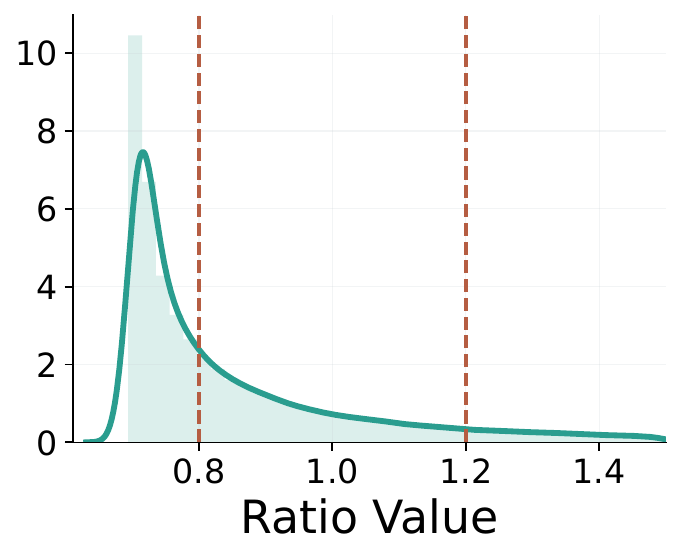}
    \captionof{figure}{Ratio distribution induced by the two policies in the LQR case study at a representative state. Dashed lines mark \(1\pm\kappa\) with \(\kappa=0.2\).}
    \label{fig:motivation_lqr_optimal_ratio}
\end{wrapfigure}

\paragraph{Scalar threshold is not a reliability criterion.} 
To clarify this distinction,  Figure~\ref{fig:motivation_lqr_optimal_ratio} illustrates the 
ratio distribution induced by the locally optimal fixed-variance Gaussian update within the \emph{Linear-Quadratic Regulator (LQR)} case study (the construction of these two policies is detailed in Section~\ref{sec:case_study}). Most realized ratios in the figure lie outside the clipping interval $[0.8, 1.2]$. Nevertheless, as demonstrated in Section~\ref{sec:case_study}, the resulting importance-weighted estimator remains statistically reliable.  

Therefore, being outside the clipping interval $[1-\kappa, 1+\kappa]$ is not, by itself, evidence of statistical unreliability. The critical question, rather, is whether extreme ratios (i.e., the {\it tails} of the distribution) occur with sufficient frequency to dominate the empirical average. This shift redirects our focus from pointwise clipping to the comprehensive tail behavior of the likelihood ratio distribution. %

\subsection{Importance Ratio and Generalized Pareto Distribution}
\label{subsec:GPD}
The previous example shows that scalar clipping is not, in itself, a criterion for reliability. A fixed scalar threshold provides only a pointwise partition of the observed ratios and does not characterize how frequently exceedances occur or how rapidly their probabilities decay. For an importance weighted mean, the crucial question is instead whether the ratio distribution possesses sufficiently light tails to ensure the stability of the empirical average. Specifically, the existence and magnitude of tail moments dictate the convergence behavior of importance-weighted estimators \citep{vehtari2017practical,yao2018yes,vehtari2024pareto}; thus, statistical reliability is fundamentally governed by the tail behavior of the importance ratio $r$.

This observation naturally leads to the concept of threshold exceedances. For a high threshold $u$, the conditional distribution of large ratios can be studied using the exceedance law of $r-u \mid r>u$. Extreme Value Theory provides a canonical approximation for this object: under standard regularity conditions, the Pickands-Balkema-de-Haan theorem implies that the distribution of threshold exceedances is asymptotically well approximated by a Generalized Pareto Distribution (GPD, \citep{pickands1975statistical,vehtari2024pareto}). We write this approximation as:
\begin{equation}
\Pr(r-u \le y \mid r>u)
\approx
G_{k,\sigma}(y)
=
1-
\left(
1+k\frac{y}{\sigma}
\right)^{-1/k},
\qquad
y\ge 0,\;\;1+k y/\sigma>0 .
\label{eq:gpd_exceedance}
\end{equation}
The shape parameter $k$ summarizes the heaviness of the upper tail. Smaller values of $k$ correspond to lighter tails and a more stable importance-weighted average, whereas larger values correspond to heavier tails and slower, less reliable convergence. This connection forms the basis of Pareto-Smoothed Importance Sampling (PSIS). PSIS operates by fitting a GPD to the largest observed ratios and then replacing the most extreme weights with smoothed order statistics derived from the fitted tail model \citep{vehtari2017practical,yao2018yes,vehtari2024pareto}. In this role, the estimated shape parameter $k$ functions both as a diagnostic for the reliability of the importance-weighted estimator and as a mechanism for post-hoc stabilization. This concept has been particularly influential in statistics, where the ratios are typically analyzed after an importance-weighted procedure has already been completed \citep{vehtari2017practical,yao2018yes,vehtari2024pareto,cho2025fast}.

Nevertheless, policy optimization presents a crucial difference. The ratio distribution is not merely observed after the fact; rather, it is actively created by the policy update process ($\bthetaold\mapsto\btheta$). Consequently, PSIS-style smoothing does not inherently provide a differentiable objective suitable for selecting the new policy. The challenge shifts from merely diagnosing whether a given set of ratios possesses a reliable tail to designing an update whose induced ratio distribution has controllable extremes. This necessity motivates our proposed method: instead of performing post-hoc fitting and smoothing of ratios, we use the order-statistic structure of the generalized Pareto distribution to construct a proximal policy objective that proactively controls the tail behavior during the update itself.

\section{Analysis of the Tail-Controlled Regime}
\label{sec:lqr}
We now formalize the critical role of the tail parameter $k$ in determining the statistical reliability of importance-weighted policy improvement. Our central argument is that the stability of the empirical surrogate is determined not by whether individual likelihood ratios fall within a fixed scalar interval, but by the tail behavior of the likelihood ratio distribution. We utilize an LQR example to provide a concrete contrast between this distributional criterion and traditional PPO-style scalar clipping.

\subsection{Benefits of Tail Control}

We first isolate the question at the estimator level. Suppose the data used for policy improvement are independent draws from the old-policy state-action distribution, $(s_t,a_t)\stackrel{\mathrm{i.i.d.}}{\sim}d_{\bthetaold}(s)\pi_{\bthetaold}(a\mid s),$ where $d_{\bthetaold}$ denotes the state distribution induced by policy $\pi_{\btheta_\text{old}}$. For a candidate policy $\pi_{\btheta}$, we define the associated advantage-weighted random variable $X_t(\btheta)=r_t(\btheta)A_t.$ The population surrogate is therefore $\mathcal L(\btheta):=\E\{X_t(\btheta)\},$ and its empirical analogue is $\widehat{\mathcal L}_n(\btheta)= 1/n\sum_{t=1}^n X_t(\btheta)$, which serves as the empirical objective approximating the policy-improvement target $\mathcal L(\btheta)$.

\paragraph{Tail-controlled regimes.}
For each policy parameter $\btheta$, let $k(\btheta)$ denote the upper-tail index of the random variable $r_t(\btheta)$ under the measure $d_{\bthetaold}\pi_{\bthetaold}$. Specifically, we consider the regularly varying regime where $\Pr\{r_t(\btheta)>x\}=x^{-1/k(\btheta)}L_{\btheta}(x)$, and $L_{\btheta}$ is a slowly varying function. In this representation, smaller values of $k(\btheta)$ indicate faster tail decay, while larger values correspond to heavier likelihood ratio tails. Throughout this section, we assume the existence of a constant $\eta\in(0,{k(\btheta)}^{-1}-m),$ with $m>1$, such that $\E\lbrace |A_t|^{m(m+\eta)/\eta} \rbrace$ is finite. Consequently, by Hölder's inequality, this implies that $\E\left[|X_t|^m\right]$ is also finite. Under the assumption of regular variation, the tail index $k(\btheta)$ guarantees that the likelihood ratio has finite moments for all orders $p<1/k(\btheta)$ \citep{vehtari2017practical, yao2018yes, vehtari2024pareto}. To formalize this stability condition, we define the tail-controlled region for $m>1$:
\begin{equation}
\Theta_m:=\left\{\btheta:k(\btheta)<1/m\right\}.\label{eq:theta_m}
\end{equation}
Membership in $\Theta_m$ guarantees the existence of the $m$-th moment of the likelihood ratio \citep{pickands1975statistical, vehtari2024pareto}, and thus guarantees the existence of the $m$-th moment of the advantage-weighted variable $X_t(\btheta)$.
\begin{theorem}[Tail control and finite-sample estimation]\label{thm:tail_control_finite_sample}
For any parameter $\btheta$ and any threshold $u>0$, the following statements hold, with $C<\infty$ being a universal constant: 
\begin{itemize}[nosep]
    \item[(i)] if $\btheta\in\Theta_2$, then $\Pr\{\|\widehat{\mathcal L}_n(\btheta)-\mathcal L(\btheta)\|>u\}\le1/(nu^2){\Var\{X_t(\btheta)\}}$
    \item[(ii)]  if $\btheta\in\Theta_3$,  then $\Pr\{\|\widehat{\mathcal L}_n(\btheta)-\mathcal L(\btheta)\|>u\}\le1/({n^{3/2}u^3}){C\,\E|X_t(\btheta)-\mathcal L(\btheta)|^3}$.
\end{itemize}
\end{theorem}
Theorem~\ref{thm:tail_control_finite_sample} provides an estimator-level interpretation of the tail index $k$. The condition $k(\btheta)<1/2$ ensures that the importance-weighted estimator operates within a finite-variance regime, while the stronger condition $k(\btheta)<1/3$ yields sharper fixed-deviation control. Consequently, a stronger tail decay of the ratio translates directly into sharper finite-sample control of the empirical surrogate.

\subsection{Case Study: Linear-Quadratic Regulator}
\label{sec:case_study}
We use an LQR \citep{bradtke1992reinforcement,fazel2018global} example to illustrate the theoretical concepts introduced above, which serves two purposes: (i) to demonstrate that scalar likelihood ratio clipping is a misleading criterion for reliability; and (ii) to establish that tail control yields a non-degenerate and statistically meaningful alternative. Specifically, we consider a discrete-time stochastic LQR model characterized by linear-Gaussian transition dynamics and a Gaussian linear policy class:
\begin{equation}
\pi_{\btheta}(\cdot\mid s)
=
\mathcal N(K_{\btheta}s,\sigma_{\btheta}^2),
\qquad
\sigma_{\btheta}>0.
\label{eq:lqr_old_policy}
\end{equation}
Under the standard stability assumptions, the induced state distribution is Gaussian; consequently, the value function, advantage function, and likelihood ratio possess analytical forms. This allows the tail-controlled regions $\Theta_m$ to be characterized in closed form, with detailed formulas in Appendix~\ref{app:lqr_motivation_details}.
\paragraph{Scalar thresholds are not a reliability criterion.}
A fixed clipping interval assesses each likelihood ratio in isolation:
a ratio is treated differently according to whether it lies inside or
outside $[1-\kappa,1+\kappa]$. Such a criterion does not directly account for the reliability of the resulting policy estimation.

This mismatch appears in both directions. Prior work has established that finite sample ratio clipping does not guarantee distributional control. For instance, \citet{wang2020truly} showed that sampled likelihood ratios can fall within a prescribed clipping interval while the resulting policy update induces arbitrarily large, and sometimes infinite, KL divergence. On the other
hand, as illustrated in Figure~\ref{fig:motivation_lqr_optimal_ratio}, ratios may frequently lie outside the same interval while the corresponding importance-weighted estimator remains statistically well behaved.

A natural question is whether scalar thresholding can be made distributionally meaningful by imposing the same bound uniformly over the state-action space. The following proposition shows that this strengthening is degenerate even in the Gaussian LQR class.

\begin{proposition}[Degeneracy of a Globally Scalar-Bounded Region]
\label{prop:lqr_global_clip_singleton}
For all \(0<\kappa<1\), define
\[
\Theta_{\mathrm{scalar}}(\kappa)
:=
\left\{
\btheta:
1-\kappa
\le
\frac{\pi_{\btheta}(a\mid s)}
{\pi_{\bthetaold}(a\mid s)}
\le
1+\kappa
\quad
\text{for all }(s,a)\in\mathbb R^2
\right\}.
\]
Then we have $\Theta_{\mathrm{scalar}}(\kappa)=\{\bthetaold\}$.
\end{proposition}

Proposition~\ref{prop:lqr_global_clip_singleton} exposes the opposite
failure of a fixed scalar criterion. While sample-wise enforcement is too weak to characterize the induced ratio distribution, uniform enforcement is stringent as it eliminates every nontrivial Gaussian policy update. The issue is therefore not merely the choice of the
clipping radius $\kappa$. The magnitude of an individual ratio, viewed against a fixed scalar threshold, is not the appropriate object for characterizing statistical reliability.

\paragraph{Tail-controlled regions in LQR.}

The degeneracy observed above is avoided by the tail-controlled regions. Specifically, tail control permits non-zero changes in both the feedback gain and the policy 
\begin{wrapfigure}[23]{r}{0.33\columnwidth}
\vspace{-1.0em}
\centering
\includegraphics[
    width=\linewidth,
    height=0.155\textheight,
    keepaspectratio,
    trim={0.2cm 0.4cm 0.1cm 0.4cm},
    clip
]{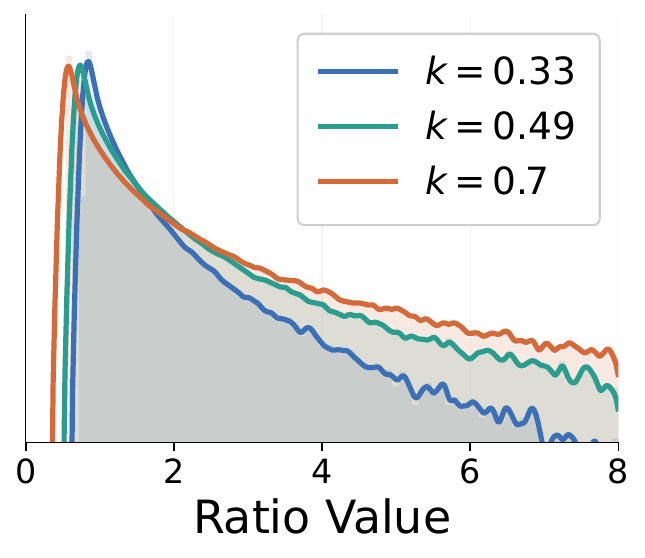}

\includegraphics[
    width=\linewidth,
    height=0.155\textheight,
    keepaspectratio,
    trim={0.5cm 0.4cm 0.1cm 0.4cm},
    clip
]{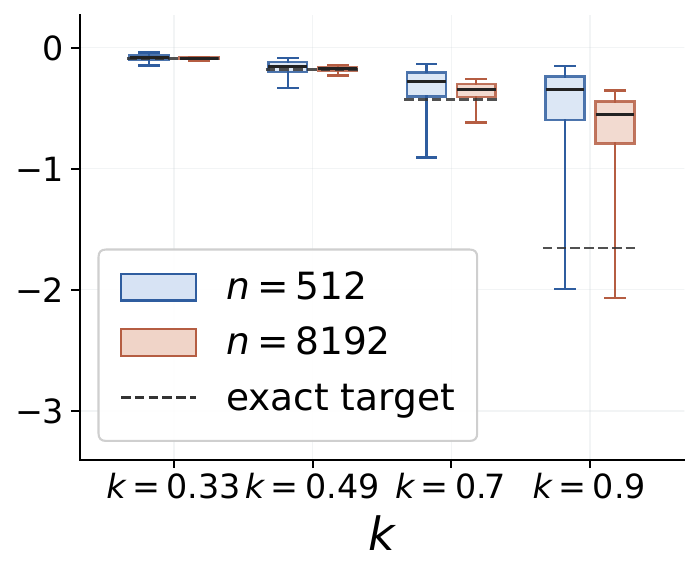}
\vspace{-0.8em}
\caption{
LQR case study. Top: heavier ratio tails emerge as \(k\) increases. Bottom: repeated estimates of \(\widehat{\mathcal L}_n\) become more dispersed as \(k\) approaches and exceeds \(1/2\).
}
\label{fig:motivation_lqr}
\end{wrapfigure}
variance while maintaining the moment regime required by Theorem~\ref{thm:tail_control_finite_sample}. Unlike scalar clipping, tail control addresses whether large ratios occur often enough to destroy the moment properties of the empirical surrogate—a question directly aligned with the statistical reliability of $\widehat{\mathcal L}_n(\btheta).$

Figure~\ref{fig:motivation_lqr} visualizes this difference. The top panel illustrates that increasing $k$ leads to a progressively heavier right tail in the likelihood ratio distribution. The bottom panel displays the corresponding impact on the empirical policy-improvement estimator. When $k$ is well below $1/2$, the repeated estimates cluster tightly around the population target. As $k$ exceeds  $1/2$, the estimates exhibit substantially increased variability.

The ratio distribution in Figure~\ref{fig:motivation_lqr_optimal_ratio} corresponds to a candidate policy chosen near the boundary of $\Theta_2$, specifically $k=0.49$. Although most realized ratios in Figure~\ref{fig:motivation_lqr_optimal_ratio} lie outside the standard PPO clipping interval $[0.8, 1.2]$, the estimator remains securely within the finite-variance regime. This demonstrates that falling outside a scalar clipping interval is not, by itself, evidence of statistical unreliability.

The remaining challenge is how to ensure this reliable regime. We address this by leveraging the Generalized Pareto approximation introduced in the preliminaries: rather than assigning a uniform radius to all ratios, we prescribe a desired profile for the selected threshold exceedances.

\section{The Proposed Algorithm: Ordered Policy Optimization}
\label{sec:method}

\begin{figure}[!htpb]
    \centering
    \includegraphics[width=0.85\linewidth,
        trim={2cm 3.5cm 2cm 1cm},
        clip]{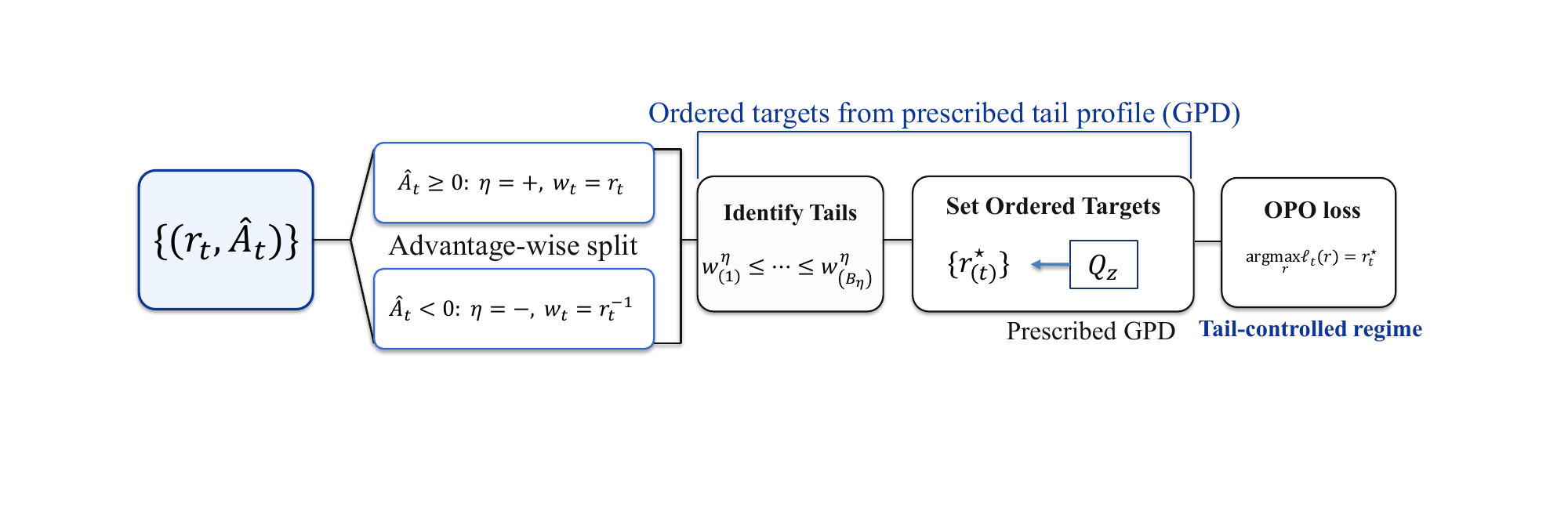}
    \caption{ Overview of the \textsc{OPO} algorithm. OPO generates the target $r_t^{\star}$ from the GPD teacher profile, then assigns each target to the corresponding order statistic: the $t$-th ordered ratio $r_{(t)}$ is assigned the corresponding $r_t^{\star}$. This process encourages the empirical tail ratios to follow the prescribed shape.}
    \label{fig:overview_algorithm}
\end{figure}
In this section, we introduce \textsc{OPO}, a policy optimization rule that replaces the scalar clipping mechanism with an order-statistic approach applied to the empirical likelihood ratio tails. The preceding analysis motivates using a GPD to model and control the tail distribution. OPO instantiates this principle using a \emph{GPD teacher profile}, whose quantiles provide structured targets across the ordered extremes. In contrast to a fixed scalar radius, the permitted ratio magnitude varies systematically with its position in the empirical tail.

Intuitively, \textsc{OPO} optimizes the policy by nudging the selected extreme ratios toward the ordered targets specified by a GPD teacher profile. As these ratios move toward the prescribed profile, the empirical tail is encouraged toward the tail regime characterized by the target shape parameter $k^\star$. The GPD provides a natural teacher for this purpose because threshold exceedances admit an asymptotic GPD representation under the classical PBdH framework (Section~\ref{subsec:GPD}). Further discussion of this modeling choice and empirical diagnostics of the observed ratio tails is provided in Appendix~\ref{app:details_for_GPD}.

To implement this idea, \textsc{OPO} accounts for the direction favored by the policy objective. Since the single sample surrogate contribution is $r_t(\btheta)\hat A_t$, positive advantages favor large ratios, whereas negative advantages favor small ratios. We therefore control the upper tail of $r_t(\btheta)$ for positive advantage samples and the upper tail of $r_t(\btheta)^{-1}$ for negative advantage samples. Further, the GPD profile is applied only to the selected threshold exceedances, comprising $M_+$ positive advantage samples and $M_-$ negative advantage samples. All remaining samples retain the nominal radius $\kappa$.

Overall, the \textsc{OPO} algorithm consists of three steps: identifying the tail partitions, assigning the selected order statistics their GPD teacher targets, and converting these targets into sample wise proximal radii. We explain these in turn.

\subsection{Advantage-aware tail partition}
\label{subsec:tail_partition}
We first identify the samples to which the teacher profile is applied. Positive-advantage samples are ordered based on the likelihood ratio \(r_t(\theta)\), while negative-advantage samples are ordered based on the inverse ratio \(r_t(\theta)^{-1}\). This ensures that both extremes are consistently represented as upper tails. Let \(\{r_t(\btheta)\}_{t=1}^{B}\) denote the likelihood ratios in a minibatch of size \(B\). We define 
\[
\mathcal I_+=\{t:\hat A_t\ge 0\},
\qquad
\mathcal I_-=\{t:\hat A_t<0\},
\qquad
B_\pm=|\mathcal I_\pm|
\]
to be the signed index sets. For the positive-advantage set, we use the ratios directly; for the negative-advantage set, we use the inverse ratios:
\[
w_t^+ = r_t(\btheta), \quad t\in\mathcal I_+,
\qquad
w_t^- = r_t(\btheta)^{-1}, \quad t\in\mathcal I_-.
\]
For each sign \(\eta\in\{+,-\}\), we sort the corresponding scores such that $w^\eta_{(1)}\le \cdots \le w^\eta_{(B_\eta)}$, with associated indices \(\tau^\eta(1),\ldots,\tau^\eta(B_\eta)\). The largest ordered scores in each group constitute the advantage-compatible empirical tails.
Consistent with the generalized Pareto approximation, which models the distribution of exceedances above a high threshold rather than the bulk of the empirical distribution, we only retain samples that represent these threshold exceedances in each signed region. Accordingly, we define the number of retained extremes as $M_\pm=\lfloor \min(0.2B_\pm,3\sqrt{B_\pm})\rfloor$, 
adopting the threshold selection rule established by \citet{vehtari2024pareto}. We then define $\mathcal T_+$ and $\mathcal T_-$ as the index sets corresponding to the largest $M_+$ and $M_-$ ordered signed scores in the respective groups. Consequently, $\mathcal T_+$ contains the samples with the largest positive-advantage ratios, while $\mathcal T_-$ contains the samples with the smallest ratios among the negative-advantage set.

\subsection{Teacher targets from tail order statistics}
\label{subsec:teacher_dist}
We next assign target values to the selected ordered extremes. Instead of fitting a GPD within each minibatch, we utilize a fixed GPD quantile curve as a \emph{teacher profile}. This approach preserves the desired tail geometry while avoiding the instability inherent in minibatch-level tail estimation.

For $z=1,\ldots,M$, we set the normalized quantile parameter $p_z=(z-1/2)/M$. Given a threshold $u$, teacher shape $k^\star$, and teacher scale $\sigma^\star$, we define the $p_z$-quantile of the teacher GPD:
\[
Q_z(u,M)
=
u+{\sigma^\star}/{k^\star}
\{(1-p_z)^{-k^\star}-1\},
\]
which corresponds to the inverse-CDF mapping of the GPD. We apply this profile independently to each signed tail. For $\eta\in\{+,-\}$, the empirical threshold $u_\eta$ is set as the boundary point. 
The $z$-th selected order statistic in the $\eta$-tail is then assigned the teacher value $Q_z(u_\eta,M_\eta)$. Expressing these targets directly in the likelihood ratio space, we obtain:
\[
r_{\tau^\eta{(B_\eta-M_\eta+z)}}^{\eta,\star}
=
\begin{cases}
Q_z(u_\eta,M_\eta), & \eta=+,\\
\{Q_z(u_\eta,M_\eta)\}^{-1}, & \eta=-,
\end{cases}
\qquad
z=1,\ldots,M_\eta .
\]
These targets effectively define sample-wise radii along the advantage-favored direction while imposing a common, ordered profile on the selected empirical tails.

\subsection{A practical proximal surrogate}
\label{subsec:surrogate_obj}
We now convert the teacher targets into a policy objective. For a single sample where \(\hat A_t\neq 0\), we consider the following constrained optimization problem:
\begin{equation}
\label{eq:eps_rule_piecewise}
\max_{r>0}\; r\hat A_t
\qquad
\text{subject to}
\qquad
|r-1|\le \varepsilon_t , \qquad
\varepsilon_t=
\begin{cases}
r_t^{+,\star}-1, & t\in\mathcal T_+,\\
1-r_t^{-,\star}, & t\in\mathcal T_-,\\
\kappa, & \text{otherwise}.
\end{cases}
\end{equation}
Here, \(\varepsilon_t\) represents the sample-wise radius induced by the assigned targets. This constrained problem is solved analytically by $r^\star=1+\varepsilon_t\operatorname{sign}(\hat A_t)$. Following the approach of \citet{xie2024simple}, we replace this hard constraint with a quadratic penalty term, which preserves the same single-sample maximizer. This yields the objective function:
\[
\mathcal{L}_{\mathrm{OPO}}(\btheta)
=
\E
\big[
r_t(\btheta)\hat A_t
-
{|\hat A_t|}/{(2\varepsilon_t)}
\{r_t(\btheta)-1\}^2
\big].
\]
This objective maintains the form of a simple quadratic policy loss, where the effective radius depends directly on the ordered empirical tail location of each sample. At the single-sample level, the maximizer is $1+\varepsilon_t\operatorname{sign}(\hat A_t)$, meaning that the selected tail samples are assigned optima consistent with the teacher profile. This structure keeps the objective simple to optimize while explicitly encoding the order-based tail control that motivates the method. The complete algorithm is detailed in Algorithm~\ref{alg:opo}. Apart from the construction of the sample-wise radius and the substitution of PPO's policy loss, all other components of the optimization pipeline remain identical to PPO. Therefore, \textsc{OPO} achieves refinement in its update geometry through tail-adaptive proximal control, all while preserving the implementation simplicity of PPO.

\section{Empirical Evaluation}
\label{sec:experiments}
We evaluate whether order-based control of likelihood ratio tails leads to more reliable policy optimization by conducting experiments on continuous-control tasks from MuJoCo-v4 \citep{todorov2012mujoco} and on Atari benchmarks \citep{bellemare2013arcade}. In these experiments, \textsc{OPO} uses fixed hyperparameters across all tasks: $\kappa=0.2,~\sigma^\star=0.1,~k^\star=0.33$. Note that the choice of $k^\star=0.33$ falls within the reliable tail-moment regime motivated by Theorem~\ref{thm:tail_control_finite_sample}. All compared methods use the same learning rate, rollout length, and minibatch size.

\subsection{Experiments on MuJoCo-v4}
\label{subsec:mujoco}
We compare \textsc{OPO} against several established methods: clipped Proximal Policy Optimization (PPO \citep{schulman2017proximal}), Proximal Policy Optimization with the KL penalty (PPO-KL \citep{schulman2017proximal}), Trust Region-Guided Proximal Policy Optimization \citep{wang2019trust}, Simple Policy Optimization (SPO \citep{xie2024simple}), and Proximal Policy Optimization with adaptive clipping (PPO-$\lambda$ \citep{chen2018adaptive}). We test these methods on six standard MuJoCo tasks: Ant, HalfCheetah, Hopper, Humanoid, HumanoidStandup, and Walker2d, running each experiment over five independent seeds.

We follow the evaluation protocol established by \citet{xie2024simple}. First, to mitigate sensitivity to policy-network capacity, we evaluate each method using two different policy parameterizations: a three-layer network and a seven-layer network. For each method, we calculate the average validation return over the final evaluation window, which corresponds to the last 10\% of the total training steps. We report the better of the two architecture-specific average returns for each method and environment. Consistent with the normalization protocol of \citet{xie2024simple}, all scores in each environment are normalized using the range of validation returns attained by PPO-Clip during training, defined as $\mathrm{NormScore}={(\mathrm{score}-\mathrm{min}_{\mathrm{PPO}})}/{(\mathrm{max}_{\mathrm{PPO}}-\mathrm{min}_{\mathrm{PPO}})}$, where $\mathrm{max}_{\mathrm{PPO}}$ and $\mathrm{min}_{\mathrm{PPO}}$ denote the maximum and minimum validation returns achieved by PPO on that specific environment.

\begin{figure}[h]
    \centering    
    \includegraphics[width=0.9\linewidth]{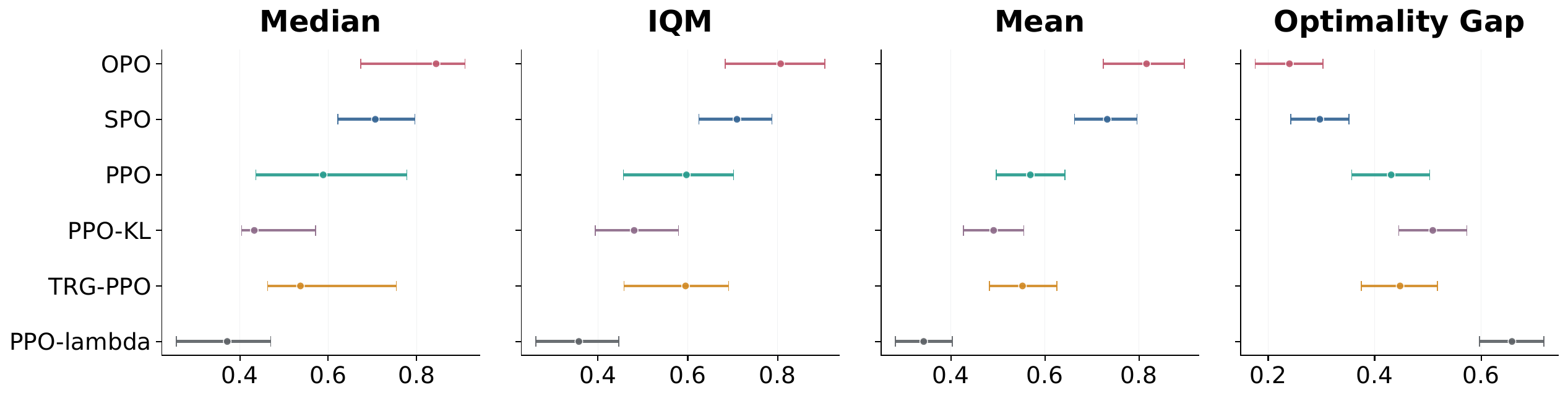}
    \caption{
    Aggregate MuJoCo performance under the architecture-selected
    protocol. We report the median, interquartile mean (IQM),
    mean, and optimality gap across tasks. Higher is better for the
    first three metrics, and lower is better for the optimality gap. 
    }
    \label{fig:mujoco_aggregate}
\end{figure}

\textbf{Results.}
Our results are summarized in Figure~\ref{fig:mujoco_aggregate}, which aggregates the normalized scores using the median, interquartile mean (IQM), mean, and optimality gap calculated via stratified bootstrapping, as suggested in \cite{agarwal2021deep}. Higher values indicate better performance for the median, IQM, and mean, while lower values are desirable for the optimality gap. Across these robust aggregate statistics, \textsc{OPO} is among the strongest methods and achieves a small optimality gap.

\begin{figure}[!htbp]
    \centering
    \vspace{-0.2in}
    \includegraphics[width=0.9\linewidth]{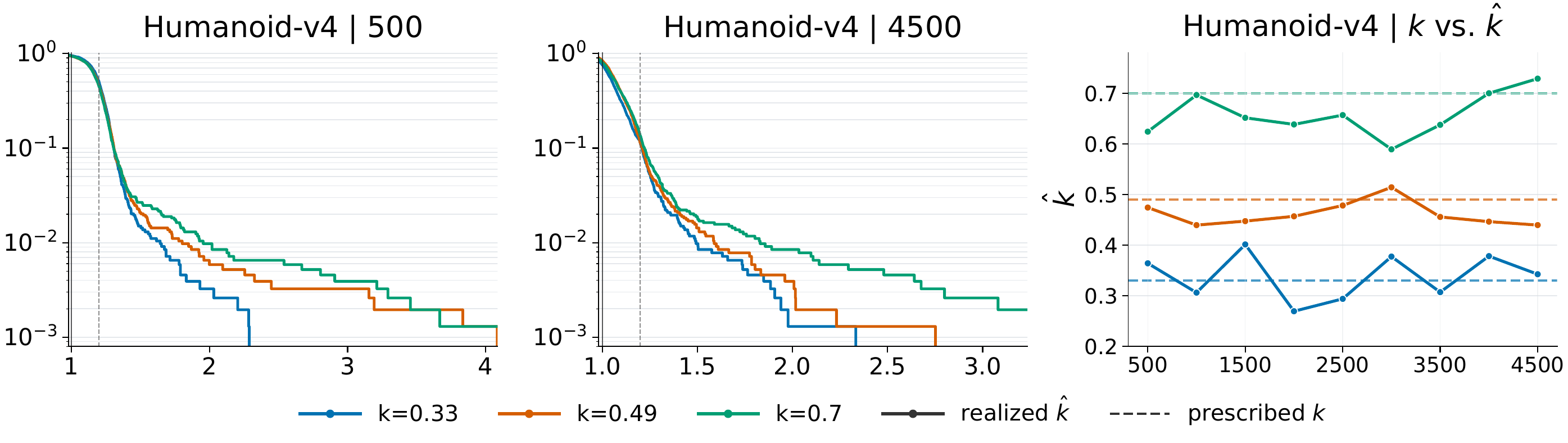}
    \caption{
    Effect of the generalized Pareto shape parameter on the
    empirical upper tail of the policy ratio. Left and Middle: At a representative time step, we plot the empirical survival function
    \(\widehat{\bar F}(x)=B^{-1}\sum_{t=1}^{B}\mathbf{1}\{r_t>x\}\)
    of the raw policy ratio
    \(r_t\)
    for three prescribed GPD shape parameters $k\in\{0.33,0.49,0.70\}$. Larger \(k\) leads to slower tail
    decay. Right: Evolution of the estimated tail shape $\hat{k}$ across updates, with dashed lines indicating the corresponding prescribed values of $k$. %
    }
    \label{fig:ratio_tail_survival_shape}
\end{figure}

\textbf{Effect of the Shape Parameter.}
We next investigate how the generalized Pareto shape parameter affects the tail geometry of the policy update. Figure~\ref{fig:ratio_tail_survival_shape} displays the empirical survival function of the policy ratios, $\widehat{\bar F}(x) = B^{-1}\sum_{t=1}^{B}\mathbf{1}\{r_t>x\}$, evaluated at a representative optimization point for three prescribed GPD shape parameters: $k\in\{0.33,0.49,0.70\}$. As anticipated, the curves illustrate that increasing $k$ induces slower tail decay. The right panel further shows that the estimated tail shape $\hat{k}$ broadly tracks the prescribed value throughout training. Furthermore, we show in Appendix~\ref{app:mujoco-further-results} that $k=0.33$ generally yields stronger performance,  consistent with our motivating theoretical analysis.
\begin{figure}[!htbp]
    \centering
    \includegraphics[width=0.89\linewidth,
        trim={0.3cm 0.4cm 0.5cm 0.2cm},
        clip]{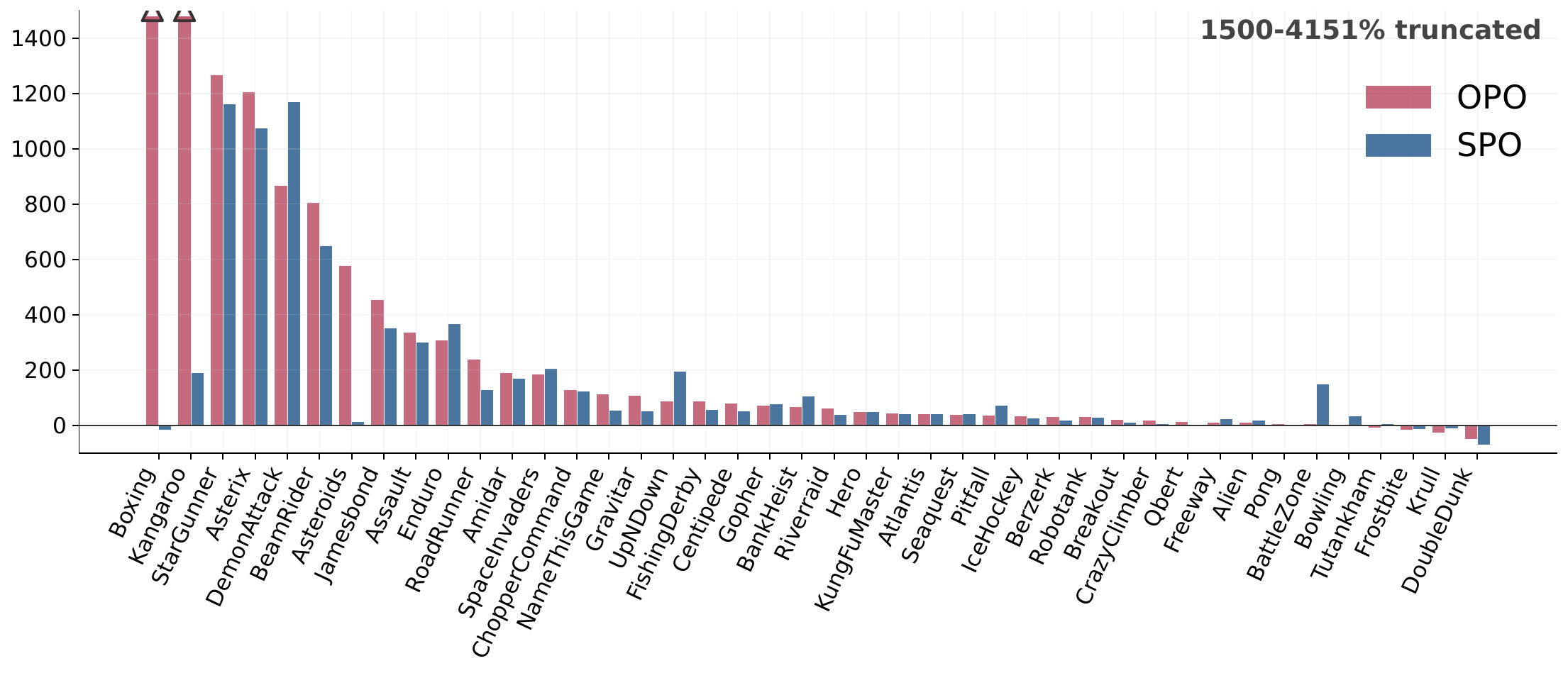}
    \caption{
        Relative improvement over PPO on Atari environments. Each bar reports the relative improvement observed over the final 10\% of validation runs for \textsc{OPO} and SPO. Bars exceeding the visible range are truncated for readability, and this truncation is indicated in the plot.
    }
    \label{fig:atari_relative_improvement}
\end{figure}

\subsection{Experiments on Atari}
\label{subsec:atari}

We further evaluate \textsc{OPO} on Atari to serve as a discrete-control stress test. Adopting the evaluation protocol from \citet{xie2024simple}, we report the relative improvement of \textsc{OPO} over PPO. We conduct these experiments using three different random seeds and utilizing ResNet-18 \citep{he2016deep} as the encoder.

\textbf{Results.}
As shown in Figure~\ref{fig:atari_relative_improvement}, \textsc{OPO} achieves competitive relative gains across the Atari games. Notably, it often matches or exceeds the performance of SPO while retaining the simple first-order optimization structure. This suggests that \textsc{OPO} is not restricted to continuous control settings but can successfully transfer its benefits to discrete-action policy optimization.

\begin{figure}[h]
    \centering
    \includegraphics[width=0.32\linewidth]{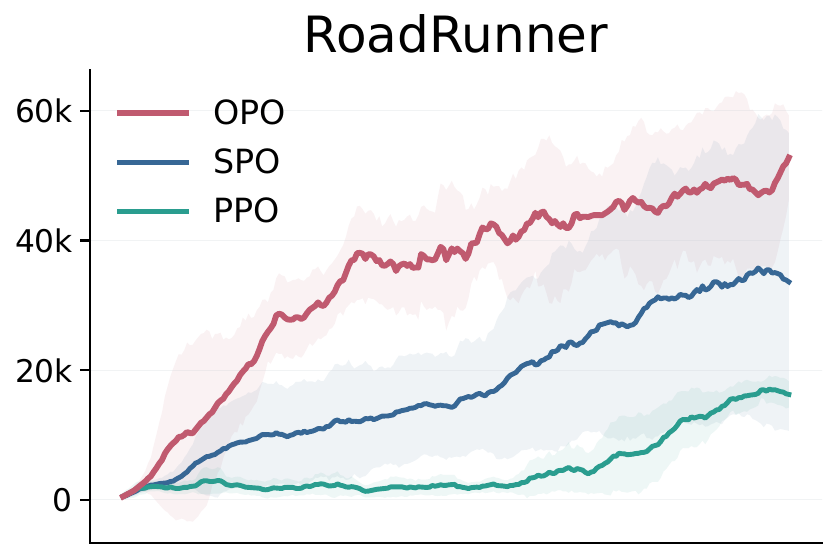}
    \includegraphics[width=0.32\linewidth]{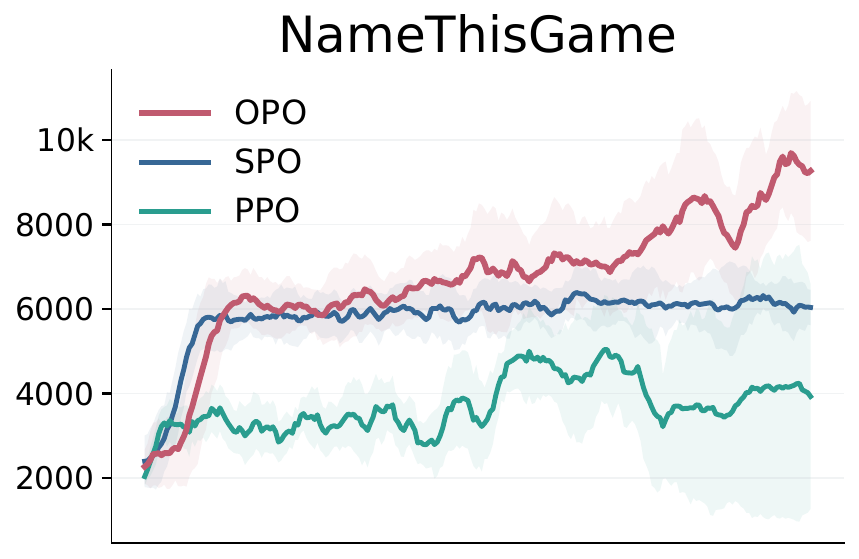}
    \includegraphics[width=0.32\linewidth]{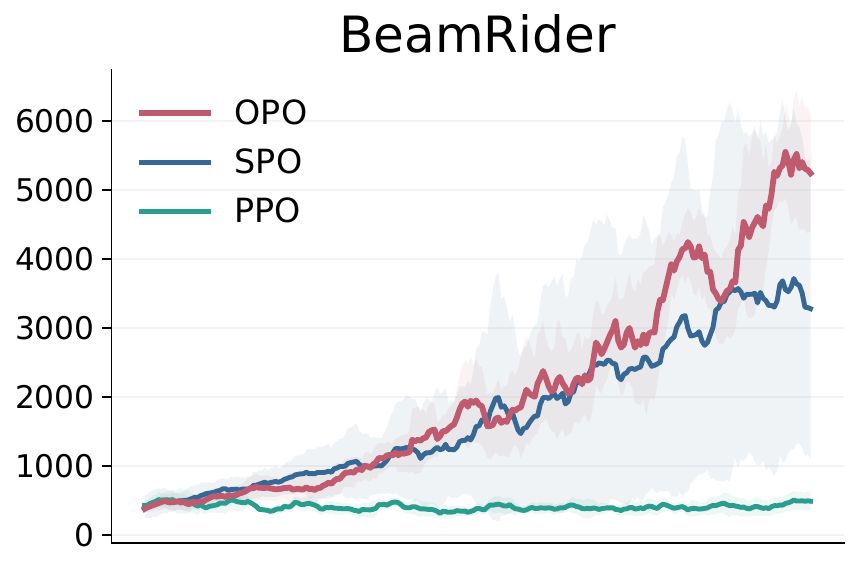}
    \caption{
    Representative Atari learning curves. Left: RoadRunner. Middle: NameThisGame. Right:
    BeamRider. \textsc{OPO} reaches higher returns than PPO and SPO on all tasks.
    }
    \label{fig:atari_learning_curves}
\end{figure}

Figure~\ref{fig:atari_learning_curves} provides three representative
learning curves. On RoadRunner, NameThisGame, and BeamRider, \textsc{OPO} reaches higher final returns than the PPO and SPO
baselines. Consistent with the central empirical
claim, controlling the ordered tails of the likelihood ratio distribution yields a competitive performance while retaining the implementation simplicity of PPO. Additional results, including learning curves, are provided in Appendix~\ref{app:atari_further_results}.

\section{Conclusion}
We addressed a limitation of PPO in stabilizing policy updates. PPO's reliance on a fixed, pointwise clipping range, while offering computational simplicity, does not directly account for the tail behavior governing the reliability of the importance-weighted surrogate. We demonstrated that effective policy stabilization can benefit from controlling the tail behavior of the likelihood ratio distribution, rather than merely enforcing local constraints. We proposed Ordered Policy Optimization (\textsc{OPO}). \textsc{OPO} replaces fixed pointwise clipping with an order-statistic-based approach: by identifying the influential empirical tails of the importance ratios, we assign adaptive, sample-wise proximal radii derived from a prescribed teacher profile. This modification allows the resulting optimization objective to remain first-order and highly practical, requiring only standard minibatch sorting and a minor change to the proximal penalty. Our empirical evaluations across standard control benchmarks, including MuJoCo and Atari, support the effectiveness of this approach, showing that \textsc{OPO} yields strong performance relative to established baselines while preserving the implementation simplicity of PPO. Future research directions include developing data-adaptive tail targets, moving beyond fixed teacher profiles, and deriving sharper theoretical guarantees regarding convergence and stability under function approximation.

\section*{Acknowledgments}
The authors would like to thank the anonymous reviewers for their constructive comments and valuable feedback. DC is supported in part by the National Science Foundation under Grant No. DMS-2346292 and the National Institutes of Health under Grant No. 5R01CA280970-04. PX is supported in part by the National Science Foundation under Grant Nos. DMS-2323112 and IIS-2541960 (CAREER), and by the Whitehead Scholars Program at the Duke University School of Medicine. Any opinions, findings, and conclusions or recommendations expressed in this material are those of the authors and do not necessarily reflect the views of any funding agency.

\medskip
\bibliographystyle{plainnat}
\bibliography{references}

\appendix

\section{Details for the GPD}
\label{app:details_for_GPD}

\subsection{The role of GPD in OPO}

The generalized Pareto distribution (GPD) is used in OPO as a model for the extreme tail of the likelihood ratio distribution, rather than as a parametric model for the full distribution of policy ratios. This choice is motivated by the aforementioned classical Pickands-Balkema-de Haan theorem (PBdH), under which exceedances above a sufficiently high threshold admit an asymptotic GPD approximation under standard regularity conditions. Accordingly, the role of the GPD in OPO is to provide a principled representation of extreme ratios once the analysis is restricted to the tail.

Hence, the OPO construction does not require the complete likelihood ratio distribution to follow a heavy tailed parametric family. The OPO algorithm works well when the tail observations can be meaningfully represented by a GPD approximation. 

\subsection{Investigation of PBdH}

\paragraph{Assumptions for PBdH}
Specifically, PBdH requires only the maximum domain of attraction (MDA) condition, and is not limited to a specific parametric assumption. Let $F(x)$ be the distribution of $R_1, ..., R_n$. We say that $F
\in
\operatorname{MDA}(G)$ when, for samples $(R_1,\ldots,R_n)$ from $F$, there exist $a_n>0$ and $b_n\in\mathbb R$ such that

$$
\frac{\max_{1\le i\le n}R_i-b_n}{a_n}
\Rightarrow
G.
$$
While it is hard to verify this condition in practice, this standard condition is satisfied by most nondegenerate distributions and widely accepted as mild in statistical literature\citep{balkema1974residual, pickands1975statistical, bobbia2021coupling, de2006extreme,vehtari2024pareto}. Further, PBdH imposes no parametric model on the transition dynamics, rewards, or the body of the ratio distribution, while permitting all three classical extreme-value regimes.

\paragraph{Empirical Validations of GPD}

\begin{figure}
    \centering
    \includegraphics[width=0.85\linewidth]{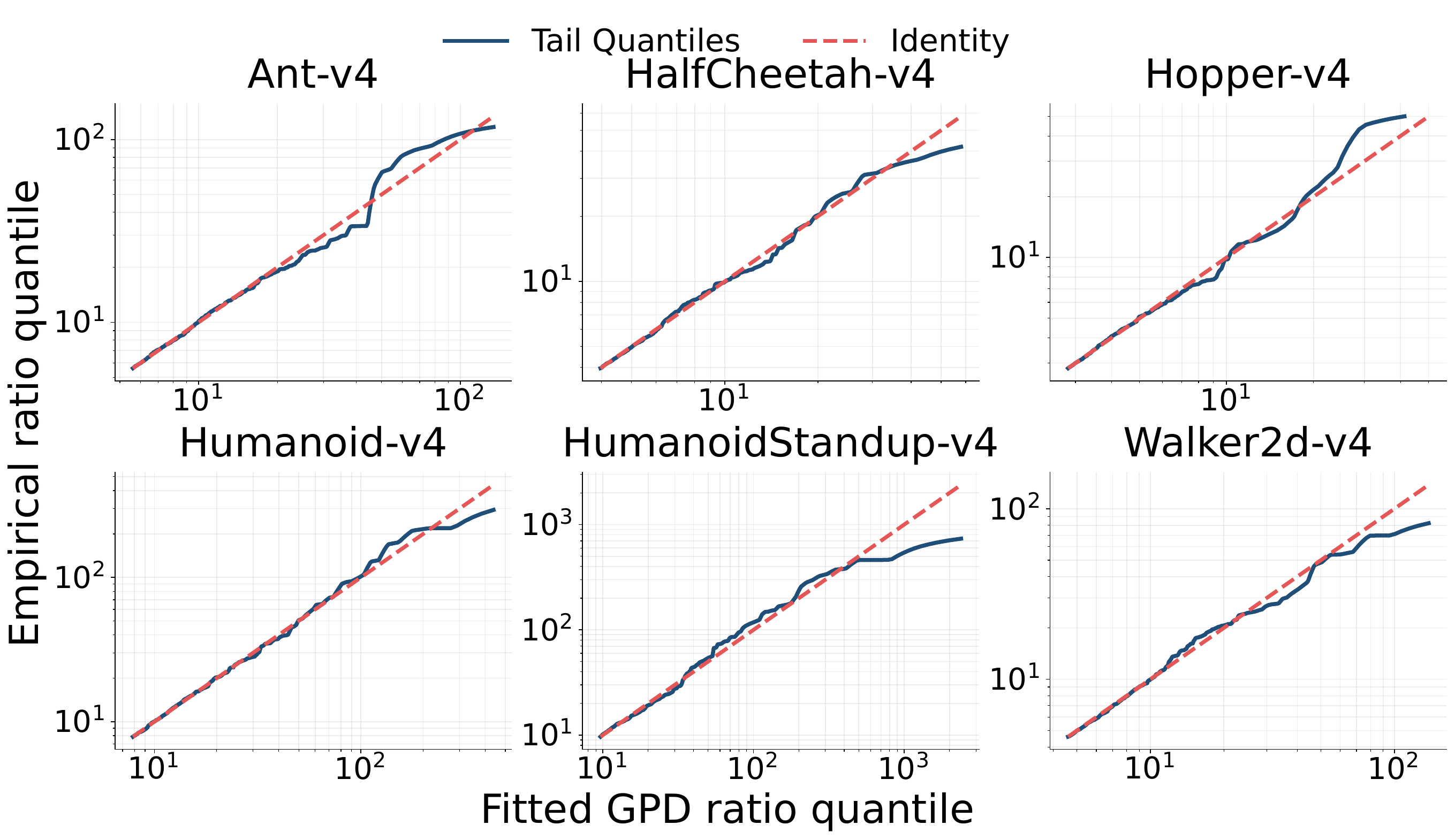}
    \caption{QQ correlation plot with the GPD.}
    \label{fig:gpd_QQ_plots}
\end{figure}

We examine this premise empirically in MuJoCo environments by fitting a GPD to the observed ratio exceedances during PPO training. Table~\ref{tab:gpd_diagnostics} reports the estimated shape parameter $\hat{k}$ together with the correlation of the corresponding GPD QQ plot in Figure~\ref{fig:gpd_QQ_plots}.

\begin{table}[h]
\centering
\caption{Estimated GPD shape parameters and QQ correlations.}
\label{tab:gpd_diagnostics}
\begin{tabular}{lcc}
\toprule
\textbf{Environment} & $\hat{k}$ & \textbf{QQ correlation} \\
\midrule
Ant              & 0.963 & 0.871 \\
HalfCheetah      & 0.465 & 0.980 \\
Hopper           & 0.522 & 0.978 \\
Humanoid         & 0.665 & 0.906 \\
HumanoidStandup  & 1.002 & 0.817 \\
Walker2d         & 0.513 & 0.961 \\
\bottomrule
\end{tabular}

\end{table}

The high GPD QQ correlations highlight the agreement between the fitted GPD tails and the observed ratios. These diagnostics indicate that the extreme likelihood ratios encountered during PPO training are well represented by a GPD tail model in a range of nonlinear control problems.

Crucially, PPO clipping does not preclude heavy-tailed likelihood-ratio distributions. Clipping modifies the contribution of extreme ratios to the surrogate objective rather than directly constraining the ratios themselves. Consequently, the raw ratio distribution may retain substantial tail mass even when the optimization objective is clipped. A similar phenomenon has been observed in prior work \citep{xie2024simple}.

\section{Details for the Analysis Section}

We provide proofs of the theoretical results in the main text, starting with the finite-sample estimation theorem. 

\subsection{Proof of Theorem~\ref{thm:tail_control_finite_sample}}
\begin{theorem}[Tail control and finite-sample estimation]
\label{thm:app_tail_control_finite_sample}
For any $\btheta$ and $u>0$, the following statements hold,
with $C<\infty$ a universal constant:
\[
\begin{aligned}
\text{\rm(i)}\quad
&\text{if } \btheta\in\Theta_2,\text{ then }
\Pr\!\left\{
\left|
\widehat{\mathcal L}_n(\btheta)-\mathcal L(\btheta)
\right|>u
\right\}
\le
\frac{\Var\{X_t(\btheta)\}}{n u^2},
\\[0.6em]
\text{\rm(ii)}\quad
&\text{if } \btheta\in\Theta_3,\text{ then }
\Pr\!\left\{
\left|
\widehat{\mathcal L}_n(\btheta)-\mathcal L(\btheta)
\right|>u
\right\}
\le
\frac{
C\,\E|X_t(\btheta)-\mathcal L(\btheta)|^3
}{
n^{3/2}u^3
}.
\end{aligned}
\]
\end{theorem}
\begin{proof}

Fix $\btheta$ throughout the proof, and write
\[
X_t := X_t(\btheta), \qquad
\mathcal L := \mathcal L(\btheta), \qquad
\widehat{\mathcal L}_n := \widehat{\mathcal L}_n(\btheta).
\]
Setting $Y_t := X_t-\mathcal L$, we have $\E[Y_t]=0$ and
\[
\widehat{\mathcal L}_n-\mathcal L
=
\frac{1}{n}\sum_{t=1}^n Y_t.
\]
We also write the second and third moments as follows:
\[
v_{\btheta}^2 := \E[Y_t^2],
\qquad
\rho_{\btheta} := \E[|Y_t|^3].
\]

Suppose first that $\btheta\in\Theta_2$. By the assumed moment condition
and H\"older's inequality,
\[
\E[|X_t|^2]
=
\E[|r_t(\btheta)A_t|^2]
<\infty,
\]
and hence $v_{\btheta}^2=\Var(X_t)<\infty$. Since
$Y_1,\ldots,Y_n$ are i.i.d. with mean zero, it follows that
\[
\E\!\left[
\left(
\widehat{\mathcal L}_n-\mathcal L
\right)^2
\right]
=
\E\!\left[
\left(
\frac{1}{n}\sum_{t=1}^n Y_t
\right)^2
\right]
=
\frac{v_{\btheta}^2}{n}.
\]
Consequently, Markov's inequality applied to
$(\widehat{\mathcal L}_n-\mathcal L)^2$ yields the following inequality for any $u>0$:
\[
\Pr\!\left(
\left|
\widehat{\mathcal L}_n-\mathcal L
\right|>u
\right)
\le
\frac{
\E\!\left[
(\widehat{\mathcal L}_n-\mathcal L)^2
\right]
}{u^2}
=
\frac{v_{\btheta}^2}{nu^2}.
\]
This proves item~(i).

Suppose next that $\btheta\in\Theta_3$. The assumed moment condition gives
$\E[|X_t|^3]<\infty$. Moreover,
\begin{align*}
|X_t-\mathcal L|^3
&\le
\bigl(|X_t|+|\mathcal L|\bigr)^3 \notag\\
&\le
4\bigl(|X_t|^3+|\mathcal L|^3\bigr),
\end{align*}
where the first step follows from the triangle inequality, and the
second from convexity of the map $x\mapsto x^3$. Combining this bound
with Jensen's inequality, $|\mathcal L|^3
=|\E[X_t]|^3\le \E[|X_t|^3]$, yields
\begin{align*}
\rho_{\btheta}
=
\E[|Y_t|^3]
&\le
4\left\{
\E[|X_t|^3]+|\mathcal L|^3
\right\} \notag\\
&\le
8\E[|X_t|^3]
<\infty.
\end{align*}

We now apply the Marcinkiewicz--Zygmund inequality with $p=3$
\citep{lubinsky1998marcinkiewicz}. Since $Y_1,\ldots,Y_n$ are
independent and mean-zero, there exists a constant $C<\infty$ such that
\[
\E\!\left[
\left|
\sum_{t=1}^n Y_t
\right|^3
\right]
\le
C\E\!\left[
\left(
\sum_{t=1}^n Y_t^2
\right)^{3/2}
\right].
\]
Using
\[
\left(\sum_{t=1}^n y_t^2\right)^{3/2}
\le
n^{1/2}\sum_{t=1}^n |y_t|^3,
\]
we obtain
\[
\E\!\left[
\left|
\sum_{t=1}^n Y_t
\right|^3
\right]
\le
C n^{1/2}\sum_{t=1}^n \E[|Y_t|^3]
=
C n^{3/2}\rho_{\btheta}.
\]
It follows that
\[
\E\!\left[
\left|
\widehat{\mathcal L}_n-\mathcal L
\right|^3
\right]
\le
\frac{C\rho_{\btheta}}{n^{3/2}}.
\]
Hence, for any $u>0$, Markov's inequality gives
\[
\Pr\!\left(
\left|
\widehat{\mathcal L}_n-\mathcal L
\right|>u
\right)
\le
\frac{
\E\!\left[
|\widehat{\mathcal L}_n-\mathcal L|^3
\right]
}{u^3}
\le
\frac{C\rho_{\btheta}}{n^{3/2}u^3},
\]
which proves item~(ii).
\end{proof}

\subsection{Proof of Proposition~\ref{prop:lqr_global_clip_singleton}}
\begin{proposition}[Consequences of Scalar Bounded Region]
For all \(0<\kappa<1\), define
\[
\Theta_{\mathrm{scalar}}(\kappa)
:=
\left\{
\btheta:
1-\kappa
\le
\frac{\pi_{\btheta}(a\mid s)}
{\pi_{\bthetaold}(a\mid s)}
\le
1+\kappa
\quad
\text{for all }(s,a)\in\mathbb R^2
\right\}.
\]
Then
\[
\Theta_{\mathrm{scalar}}(\kappa)=\{\bthetaold\}.
\]
\end{proposition}
\begin{proof}
Fix \(s\).  If \(\sigma_{\btheta}^2>\sigma_0^2\), then the ratio
\(\pi_{\btheta}(a\mid s)/\pi_{\bthetaold}(a\mid s)\) diverges as
\(|a|\to\infty\), violating the upper bound.  If
\(\sigma_{\btheta}^2<\sigma_0^2\), then the ratio converges to zero
along the action tails, violating the lower bound.  Therefore any
globally clipped policy must have
\[
\sigma_{\btheta}^2=\sigma_0^2.
\]
With equal variances, if \(K_{\btheta}\neq K_{\mathrm{old}}\), then for
any state \(s\neq0\), the two Gaussian means differ.  The density ratio
is then exponential-linear in \(a\), and hence is unbounded in one
action tail and converges to zero in the other.  This again violates
the two-sided clipping condition.  Therefore
\[
K_{\btheta}=K_{\mathrm{old}}.
\]
Thus the only globally clipped policy is the old policy itself.
\end{proof}

\section{Details for the LQR Motivating Case Study}
\label{app:lqr_motivation_details}

We first restate the LQR model. For fixed parameters
$A,B,q$, and $\lambda$, the state dynamics and reward function are given by
\begin{equation}
s_{t+1}=As_t+Ba_t+w_t,
\qquad \mathcal R(s_t,a_t)
=
-\frac{1}{2}\bigl(qs_t^2+\lambda a_t^2\bigr),
\label{eq:lqr_dynamics}
\end{equation}
where \(w_t\sim \mathcal N(0,\sigma_w^2)\), with \(q\ge 0\) and
\(\lambda>0\). Further, let the old policy be :

\[
\pi_{\bthetaold}(\cdot\mid s)
=
\mathcal N(K_{\mathrm{old}}s,\sigma_0^2).
\]
Assume also that the old closed-loop system is stable, namely
\(F_{\mathrm{old}}:=A+BK_{\mathrm{old}}\) satisfies
\(|F_{\mathrm{old}}|<1\). We will show in this section that this condition implies that
the stationary distribution \(d_{\mathrm{old}}\) exists and is
Gaussian. Under \(\pi_{\bthetaold}\), the relevant value functions are available
in closed form. 

\paragraph{Stationarity under the old policy}
We first establish the existence of a stationary state distribution. Under the Gaussian old policy,
\[
a_t=K_{\mathrm{old}}s_t+\varepsilon_t,
\qquad
\varepsilon_t\sim\mathcal N(0,\sigma_0^2),
\]
with \(\varepsilon_t\) independent of \(w_t\).  Hence the closed-loop
state recursion is
\[
s_{t+1}
=
(A+BK_{\mathrm{old}})s_t+B\varepsilon_t+w_t
=
F_{\mathrm{old}}s_t+\xi_t,
\]
where
\[
F_{\mathrm{old}}:=A+BK_{\mathrm{old}},
\qquad
\xi_t:=B\varepsilon_t+w_t.
\]
Note that \(\xi_t\) is Gaussian with mean zero and variance
\[
\sigma_\xi^2
=
B^2\sigma_0^2+\sigma_w^2,
\]
due to the basic properties of the Gaussian distribution. If \(|F_{\mathrm{old}}|<1\), then the infinite series
\[
s_\infty
=
\sum_{j=0}^{\infty}
F_{\mathrm{old}}^j \xi_{-1-j}
\]
converges in \(L^2\), since
\[
\sum_{j=0}^{\infty}
F_{\mathrm{old}}^{2j}\sigma_\xi^2
=
\frac{\sigma_\xi^2}{1-F_{\mathrm{old}}^2}
<\infty.
\]
It also defines the unique stationary solution of the scalar
autoregression. Concretely,
\begin{align*}
s_t
&=
\sum_{j=0}^{\infty}
F_{\mathrm{old}}^j \xi_{t-1-j}, \\
s_{t+1}
&=
\sum_{j=0}^{\infty}
F_{\mathrm{old}}^j \xi_{t-j} \\
&=
\xi_t
+
F_{\mathrm{old}}
\sum_{j=0}^{\infty}
F_{\mathrm{old}}^j \xi_{t-1-j} \\
&=
F_{\mathrm{old}}s_t+\xi_t.
\end{align*}
Since $\{\xi_t\}$ is i.i.d., the joint distribution of
$(s_t,s_{t+1},\ldots,s_{t+k})$ does not depend on $t$.

Since \(s_\infty\) is an \(L^2\)-limit of Gaussian
linear combinations, it is Gaussian. Writing its variance as $\sigma_s^2$, we obtain
\[
\sigma_s^2
=
F_{\mathrm{old}}^2\sigma_s^2+\sigma_\xi^2.
\]
Solving this equation gives
\begin{equation}
d_{\mathrm{old}}
=
\mathcal N(0,\sigma_s^2),
\qquad
\sigma_s^2
=
\frac{B^2\sigma_0^2+\sigma_w^2}
{1-F_{\mathrm{old}}^2}.
\label{eq:app_lqr_stationary_variance}
\end{equation}
where $d_{\mathrm{old}}$ denotes the stationary state distribution under the old policy.

\subsection{Quadratic value, action-value, and advantage}
\label{app:lqr_quadratic_structure}
Now, we derive closed-form expressions for the value, action-value, and advantage functions.

\paragraph{Derivation of the value function}
We derive $V^{\pi_{\bthetaold}}$ directly from the definition of the
discounted value, without positing its form in advance. Under the old
policy,
\[
\pi_{\bthetaold}(\cdot\mid s)
=
\mathcal N(K_{\mathrm{old}}s,\sigma_0^2),
\]
the closed-loop dynamics take the form
\[
s_{t+1}
=
(A+BK_{\mathrm{old}})s_t+B\varepsilon_t+w_t
=
F_{\mathrm{old}}s_t+\xi_t,
\]
where
\[
F_{\mathrm{old}}:=A+BK_{\mathrm{old}},
\qquad
\xi_t:=B\varepsilon_t+w_t,
\qquad
\sigma_\xi^2:=B^2\sigma_0^2+\sigma_w^2,
\]
where $\xi_t$ is mean-zero Gaussian with variance
$\sigma_\xi^2$.

We first compute the second moment of $s_t$ conditional on $s_0=s$.
Iterating the recursion gives
\[
s_t
=
F_{\mathrm{old}}^t s
+
\sum_{j=0}^{t-1}
F_{\mathrm{old}}^{\,t-1-j}\xi_j.
\]
Consequently this admits the following equation:
\begin{align*}
\E[s_t^2\mid s_0=s]
&=
\E\!\left[
\left(
F_{\mathrm{old}}^t s
+
\sum_{j=0}^{t-1}
F_{\mathrm{old}}^{\,t-1-j}\xi_j
\right)^2
\,\middle|\, s_0=s
\right] \\
&=
F_{\mathrm{old}}^{2t}s^2
+
2F_{\mathrm{old}}^t s\,
\E\!\left[
\sum_{j=0}^{t-1}
F_{\mathrm{old}}^{\,t-1-j}\xi_j
\right]
+
\E\!\left[
\left(
\sum_{j=0}^{t-1}
F_{\mathrm{old}}^{\,t-1-j}\xi_j
\right)^2
\right].
\end{align*}
Here, the middle term vanishes since $\E[\xi_j]=0$ while independence of
the innovations and $\Var(\xi_j)=\sigma_\xi^2$ give
\[
\E\!\left[
\left(
\sum_{j=0}^{t-1}
F_{\mathrm{old}}^{\,t-1-j}\xi_j
\right)^2
\right]
=
\sigma_\xi^2
\sum_{j=0}^{t-1}
F_{\mathrm{old}}^{\,2(t-1-j)}
=
\sigma_\xi^2
\sum_{j=0}^{t-1}
F_{\mathrm{old}}^{\,2j}.
\]
This leads to the following identity:
\begin{equation}
\E[s_t^2\mid s_0=s]
=
F_{\mathrm{old}}^{2t}s^2
+
\sigma_\xi^2
\sum_{j=0}^{t-1}
F_{\mathrm{old}}^{\,2j}.
\label{eq:lqr_state_second_moment_direct}
\end{equation}

Next, since
\[
a_t = K_{\mathrm{old}}s_t+\varepsilon_t,
\]
we have
\[
\mathbb E[a_t^2\mid s_0=s]
=
K_{\mathrm{old}}^2 \mathbb E[s_t^2\mid s_0=s] + \sigma_0^2,
\]
because \(\varepsilon_t\) is mean-zero and independent of \(s_t\).

Recall that the reward is
\[
\mathcal{R}(s_t,a_t)
=
-\frac12\bigl(qs_t^2+\lambda a_t^2\bigr).
\]
Thus
\begin{align*}
\mathbb E[\mathcal{R}(s_t,a_t)\mid s_0=s]
&=
-\frac12
\left(
q\,\mathbb E[s_t^2\mid s_0=s]
+
\lambda\,\mathbb E[a_t^2\mid s_0=s]
\right) \\
&=
-\frac12
\left[
(q+\lambda K_{\mathrm{old}}^2)\mathbb E[s_t^2\mid s_0=s]
+
\lambda\sigma_0^2
\right].
\end{align*}
Substituting \eqref{eq:lqr_state_second_moment_direct} gives
\begin{align*}
\mathbb E[\mathcal{R}(s_t,a_t)\mid s_0=s]
&=
-\frac12
\left[
(q+\lambda K_{\mathrm{old}}^2)
\left(
F_{\mathrm{old}}^{2t}s^2
+
\sigma_\xi^2 \sum_{j=0}^{t-1} F_{\mathrm{old}}^{\,2j}
\right)
+
\lambda\sigma_0^2
\right].
\end{align*}

By definition,
\[
V^{\pi_{\bthetaold}}(s)
=
\sum_{t=0}^\infty
\gamma^t
\mathbb E[\mathcal{R}(s_t,a_t)\mid s_0=s].
\]
Hence
\begin{align*}
V^{\pi_{\bthetaold}}(s)
&=
-\frac12
(q+\lambda K_{\mathrm{old}}^2)
\sum_{t=0}^\infty \gamma^t F_{\mathrm{old}}^{2t} s^2 \\
&\qquad
-\frac12
(q+\lambda K_{\mathrm{old}}^2)\sigma_\xi^2
\sum_{t=0}^\infty \gamma^t
\sum_{j=0}^{t-1} F_{\mathrm{old}}^{\,2j}
-\frac12\lambda\sigma_0^2\sum_{t=0}^\infty \gamma^t.
\end{align*}
The coefficient of \(s^2\) is therefore
\[
-\frac12
(q+\lambda K_{\mathrm{old}}^2)
\sum_{t=0}^\infty (\gamma F_{\mathrm{old}}^2)^t s^2
=
-\frac12
\frac{q+\lambda K_{\mathrm{old}}^2}{1-\gamma F_{\mathrm{old}}^2}
\,s^2,
\]
since the stability condition $\mid F_{\mathrm{old}}\mid <1$ and $0<\gamma<1$ imply \(\gamma F_{\mathrm{old}}^2<1\). Define $P_{\mathrm{old}}$ by
\[
P_{\mathrm{old}}
:=
\frac{q+\lambda K_{\mathrm{old}}^2}
{1-\gamma(A+BK_{\mathrm{old}})^2}.
\]

It remains to collect the constant terms.  First,
\[
\sum_{t=0}^\infty \gamma^t = \frac{1}{1-\gamma}.
\]
Second, using
\[
\sum_{j=0}^{t-1} F_{\mathrm{old}}^{\,2j}
=
\frac{1-F_{\mathrm{old}}^{2t}}{1-F_{\mathrm{old}}^2},
\]
we obtain
\begin{align*}
\sum_{t=0}^\infty \gamma^t \sum_{j=0}^{t-1} F_{\mathrm{old}}^{\,2j}
&=
\frac{1}{1-F_{\mathrm{old}}^2}
\sum_{t=0}^\infty \gamma^t (1-F_{\mathrm{old}}^{2t}) \\
&=
\frac{1}{1-F_{\mathrm{old}}^2}
\left(
\frac{1}{1-\gamma}
-
\frac{1}{1-\gamma F_{\mathrm{old}}^2}
\right) \\
&=
\frac{\gamma}{(1-\gamma)(1-\gamma F_{\mathrm{old}}^2)}.
\end{align*}
Define the constant term as 
\begin{align*}
c_V
&:=
-\frac12
(q+\lambda K_{\mathrm{old}}^2)\sigma_\xi^2
\frac{\gamma}{(1-\gamma)(1-\gamma F_{\mathrm{old}}^2)}
-\frac12\lambda\sigma_0^2\frac{1}{1-\gamma}.
\end{align*}
Since
\[
P_{\mathrm{old}}
=
\frac{q+\lambda K_{\mathrm{old}}^2}{1-\gamma F_{\mathrm{old}}^2},
\qquad
\sigma_\xi^2=B^2\sigma_0^2+\sigma_w^2,
\]
this can be written as
\[
c_V
=
-\frac{
\lambda\sigma_0^2
+
\gamma P_{\mathrm{old}}(B^2\sigma_0^2+\sigma_w^2)
}{
2(1-\gamma)
}.
\]

Combining the \(s^2\) term and the constant term, we conclude that
\[
V^{\pi_{\bthetaold}}(s)
=
c_V-\frac12 P_{\mathrm{old}}s^2,
\]
with
\[
P_{\mathrm{old}}
=
\frac{q+\lambda K_{\mathrm{old}}^2}
{1-\gamma(A+BK_{\mathrm{old}})^2},
\qquad
c_V
=
-\frac{
\lambda\sigma_0^2
+
\gamma P_{\mathrm{old}}(B^2\sigma_0^2+\sigma_w^2)
}{
2(1-\gamma)
}.
\]
\paragraph{Derivation of the action-value function}
We now derive the action-value function.  By definition,
\[
Q^{\pi_{\bthetaold}}(s,a)
=
\mathcal{R}(s,a)
+
\gamma
\mathbb E[
V^{\pi_{\bthetaold}}(As+Ba+w)
\mid s,a].
\]
Since
\[
\mathbb E[(As+Ba+w)^2\mid s,a]
=
(As+Ba)^2+\sigma_w^2,
\]
we obtain
\begin{align}
Q^{\pi_{\bthetaold}}(s,a)
&=
\gamma c_V
-\frac{\gamma}{2}P_{\mathrm{old}}\sigma_w^2
-\frac{1}{2}(qs^2+\lambda a^2)
-\frac{\gamma}{2}P_{\mathrm{old}}(As+Ba)^2
\nonumber\\
&=
\gamma c_V
-\frac{\gamma}{2}P_{\mathrm{old}}\sigma_w^2
-\frac{1}{2}
\left[
(q+\gamma A^2P_{\mathrm{old}})s^2
+
2\gamma ABP_{\mathrm{old}}sa
+
(\lambda+\gamma B^2P_{\mathrm{old}})a^2
\right].
\label{eq:app_lqr_Q_joint}
\end{align}
Define
\[
\Lambda
:=
\lambda+\gamma B^2P_{\mathrm{old}},
\qquad
M
:=
\gamma ABP_{\mathrm{old}}.
\]
Since \(\lambda>0\), we have \(\Lambda>0\).  Completing the square,
\[
\Lambda a^2+2Msa
=
\Lambda\left(a+\frac{M}{\Lambda}s\right)^2
-
\frac{M^2}{\Lambda}s^2.
\]
Thus
\[
Q^{\pi_{\bthetaold}}(s,a)
=
C_{\mathrm{old}}(s)
-
\frac{\Lambda}{2}
\bigl(a-a^\star(s)\bigr)^2,
\]
where
\begin{equation}
a^\star(s)
=
-\frac{M}{\Lambda}s
=
-\frac{\gamma ABP_{\mathrm{old}}}
{\lambda+\gamma B^2P_{\mathrm{old}}}\,s.
\label{eq:app_lqr_astar}
\end{equation}

\paragraph{Derivation of the advantage function}
Finally, since
\[
V^{\pi_{\bthetaold}}(s)
=
\mathbb E_{a'\sim\pi_{\bthetaold}(\cdot\mid s)}
[
Q^{\pi_{\bthetaold}}(s,a')
],
\]
the advantage satisfies
\begin{align}
A^{\pi_{\bthetaold}}(s,a)
&=
Q^{\pi_{\bthetaold}}(s,a)
-
V^{\pi_{\bthetaold}}(s)
\nonumber\\
&=
-\frac{\Lambda}{2}
\left[
(a-a^\star(s))^2
-
\mathbb E_{a'\sim\pi_{\bthetaold}(\cdot\mid s)}
\{(a'-a^\star(s))^2\}
\right].
\end{align}
Let
\[
\mu_{\mathrm{old}}(s):=K_{\mathrm{old}}s,
\qquad
\Delta(s):=a^\star(s)-\mu_{\mathrm{old}}(s).
\]
If \(a'\sim\pi_{\bthetaold}(\cdot\mid s)\), then
\[
a'=\mu_{\mathrm{old}}(s)+\sigma_0Z',
\qquad
Z'\sim\mathcal N(0,1).
\]
Therefore
\[
\mathbb E[(a'-a^\star(s))^2\mid s]
=
\Delta(s)^2+\sigma_0^2.
\]
Hence
\begin{equation}
A^{\pi_{\bthetaold}}(s,a)
=
-\frac{\Lambda}{2}
\left[
\lbrace a-a^\star(s)\rbrace^2-\Delta(s)^2-\sigma_0^2
\right].
\label{eq:app_lqr_advantage}
\end{equation}

\subsection{Derivation of the Moment-Controlled Region \texorpdfstring{\(\Theta_m\)}{Theta m}}
\label{app:lqr_theta_m_conditional_derivation}

We derive the explicit form of the moment-controlled region for the
scalar Gaussian policy class
\[
\pi_{\btheta}(\cdot\mid s)
=
\mathcal N(K_{\btheta}s,\sigma_{\btheta}^2),
\qquad
\sigma_{\btheta}>0,
\]
relative to the old policy
\[
\pi_{\bthetaold}(\cdot\mid s)
=
\mathcal N(K_{\mathrm{old}}s,\sigma_0^2).
\]
Under the old data-generating distribution, write
\[
s\sim d_{\mathrm{old}}=\mathcal N(0,\sigma_s^2),
\qquad
a=K_{\mathrm{old}}s+\sigma_0 Z,
\qquad
Z\sim \mathcal N(0,1),
\]
with \(Z\) independent of \(s\).  Define
\[
\Delta_K:=K_{\btheta}-K_{\mathrm{old}},
\qquad
\kappa_{\btheta}:=1-\frac{\sigma_0^2}{\sigma_{\btheta}^2},
\qquad
\eta_{\btheta}(s):=\frac{\Delta_K s}{\sigma_0}.
\]
Then
\[
a-K_{\btheta}s
=
\sigma_0 Z-\Delta_K s
=
\sigma_0\{Z-\eta_{\btheta}(s)\}.
\]
Consequently, the likelihood ratio satisfies
\begin{align}
r(s,a;\btheta)
&=
\frac{\pi_{\btheta}(a\mid s)}
     {\pi_{\bthetaold}(a\mid s)}
\nonumber\\
&=
\frac{\sigma_0}{\sigma_{\btheta}}
\exp\left\{
-\frac{(a-K_{\btheta}s)^2}{2\sigma_{\btheta}^2}
+
\frac{(a-K_{\mathrm{old}}s)^2}{2\sigma_0^2}
\right\}
\nonumber\\
&=
\sqrt{1-\kappa_{\btheta}}\,
\exp\left\{
\frac{\kappa_{\btheta}}{2}Z^2
+
(1-\kappa_{\btheta})\eta_{\btheta}(s)Z
-
\frac{1-\kappa_{\btheta}}{2}\eta_{\btheta}(s)^2
\right\}.
\label{eq:app_ratio_kappa_eta}
\end{align}

We now prove an \(m\)-moment bound for
\[
X_t(\btheta)
=
r_t(\btheta)A_t,
\qquad
A_t=A^{\pi_{\bthetaold}}(s_t,a_t).
\]
From the closed-form LQR advantage,
\[
A^{\pi_{\bthetaold}}(s,a)
=
-\frac{\Lambda}{2}
\left[
\{a-a^\star(s)\}^2-\Delta(s)^2-\sigma_0^2
\right],
\]
where
\[
\Delta(s):=a^\star(s)-K_{\mathrm{old}}s=Ds.
\]
Since \(a=K_{\mathrm{old}}s+\sigma_0 Z\), we have
\[
a-a^\star(s)=\sigma_0Z-\Delta(s).
\]
Therefore
\begin{equation}
A^{\pi_{\bthetaold}}(s,a)
=
-\frac{\Lambda}{2}
\left[
\sigma_0^2(Z^2-1)-2\sigma_0\Delta(s)Z
\right].
\label{eq:app_advantage_Z_current}
\end{equation}

Fix \(m>1\).  From \eqref{eq:app_advantage_Z_current} and the triangle
inequality,
\[
|A^{\pi_{\bthetaold}}(s,a)|
\le
\frac{\Lambda}{2}
\left\{
\sigma_0^2|Z^2-1|
+
2\sigma_0|\Delta(s)|\,|Z|
\right\}.
\]
Raising both sides to the \(m\)-th power and using
\((u+v)^m\le 2^{m-1}(u^m+v^m)\), we obtain
\[
|A^{\pi_{\bthetaold}}(s,a)|^m
\le
C_m
\left\{
|Z^2-1|^m
+
|\Delta(s)|^m|Z|^m
\right\}.
\]
Since \(\Delta(s)=Ds\), we further have
\[
|Z^2-1|^m
\le
C_m(1+|Z|^{2m}),
\qquad
|\Delta(s)|^m|Z|^m
\le
C_m(|s|^{2m}+|Z|^{2m}).
\]
Thus, after enlarging \(C_m\),
\begin{equation}
|A^{\pi_{\bthetaold}}(s,a)|^m
\le
C_m\left(1+|s|^{2m}+|Z|^{2m}\right).
\label{eq:app_advantage_polynomial_bound}
\end{equation}
Consequently,
\begin{align}
\E\!\left[
|X_t(\btheta)|^m\,\middle|\,s
\right]
&=
\E\!\left[
r(s,a;\btheta)^m
|A^{\pi_{\bthetaold}}(s,a)|^m
\,\middle|\,s
\right]
\nonumber\\
&\le
C_m(1+|s|^{2m})
\E\!\left[
(1+|Z|^{2m})r(s,a;\btheta)^m
\,\middle|\,s
\right].
\label{eq:app_reduce_to_Z_current}
\end{align}

We now bound the conditional expectation on the right-hand side.  From
\eqref{eq:app_ratio_kappa_eta},
\[
r(s,a;\btheta)^m
=
(1-\kappa_{\btheta})^{m/2}
\exp\left\{
\frac{m\kappa_{\btheta}}{2}Z^2
+
m(1-\kappa_{\btheta})\eta_{\btheta}(s)Z
-
\frac{m(1-\kappa_{\btheta})}{2}\eta_{\btheta}(s)^2
\right\}.
\]
Multiplying by the standard normal density
\(\phi(z)=(2\pi)^{-1/2}e^{-z^2/2}\), the exponent becomes
\[
-\frac{1-m\kappa_{\btheta}}{2}z^2
+
m(1-\kappa_{\btheta})\eta_{\btheta}(s)z
-
\frac{m(1-\kappa_{\btheta})}{2}\eta_{\btheta}(s)^2.
\]
Assume
\[
\kappa_{\btheta}<\frac1m.
\]
Then \(1-m\kappa_{\btheta}>0\), and completing the square yields
\begin{align*}
-\frac{1-m\kappa_{\btheta}}{2}z^2
+
m(1-\kappa_{\btheta})\eta_{\btheta}(s)z
&=
-\frac{1-m\kappa_{\btheta}}{2}
\left(
z-\mu_m(s)
\right)^2 \\
&\quad
+
\frac{m^2(1-\kappa_{\btheta})^2}{2(1-m\kappa_{\btheta})}
\eta_{\btheta}(s)^2,
\end{align*}
where
\[
\mu_m(s)
:=
\frac{m(1-\kappa_{\btheta})}{1-m\kappa_{\btheta}}
\eta_{\btheta}(s).
\]
After collecting the remaining \(\eta_{\btheta}(s)^2\) term, we get
\[
c_m(\btheta)
:=
\frac{m(m-1)(1-\kappa_{\btheta})}
{2(1-m\kappa_{\btheta})},
\]
so that
\begin{align}
\E\!\left[
(1+|Z|^{2m})r(s,a;\btheta)^m
\,\middle|\,s
\right]
&\le
C_{m,\btheta}
\exp\left\{
c_m(\btheta)\eta_{\btheta}(s)^2
\right\}
\left(1+|\mu_m(s)|^{2m}\right)
\nonumber\\
&\le
C_{m,\btheta}
(1+|s|^{2m})
\exp\left\{
c_m(\btheta)
\left(\frac{\Delta_K}{\sigma_0}\right)^2s^2
\right\}.
\label{eq:app_conditional_ratio_moment_bound}
\end{align}
Substituting \eqref{eq:app_conditional_ratio_moment_bound} into
\eqref{eq:app_reduce_to_Z_current}, we obtain
\begin{equation}
\E\!\left[
|X_t(\btheta)|^m\,\middle|\,s
\right]
\le
C_{m,\btheta}
(1+|s|^{4m})
\exp\left\{
c_m(\btheta)
\left(\frac{\Delta_K}{\sigma_0}\right)^2s^2
\right\}.
\label{eq:app_conditional_X_moment_bound}
\end{equation}

It remains to integrate over the stationary state distribution
\(s\sim\mathcal N(0,\sigma_s^2)\).  For a centered Gaussian random
variable \(S\sim\mathcal N(0,\sigma_s^2)\),
\[
\E[e^{cS^2}]<\infty
\qquad\Longleftrightarrow\qquad
2c\sigma_s^2<1.
\]
Therefore the right-hand side of
\eqref{eq:app_conditional_X_moment_bound} is integrable whenever
\[
2c_m(\btheta)
\left(\frac{\Delta_K}{\sigma_0}\right)^2
\sigma_s^2
<1.
\]
Substituting the expression for \(c_m(\btheta)\), this condition is
\begin{equation}
\frac{m(m-1)(1-\kappa_{\btheta})}
{1-m\kappa_{\btheta}}
\left(\frac{K_{\btheta}-K_{\mathrm{old}}}{\sigma_0}\right)^2
\sigma_s^2
<1.
\label{eq:app_state_integrability_current}
\end{equation}

Combining this condition with \(\kappa_{\btheta}<1/m\), we conclude
that
\[
\E[|X_t(\btheta)|^m]<\infty
\]
whenever
\[
\kappa_{\btheta}<\frac1m
\]
and
\[
|K_{\btheta}-K_{\mathrm{old}}|
<
\frac{\sigma_0}{\sigma_s}
\sqrt{
\frac{1-m\kappa_{\btheta}}
{m(m-1)(1-\kappa_{\btheta})}
}.
\]
Finally, recalling that
\[
\kappa_{\btheta}
=
1-\frac{\sigma_0^2}{\sigma_{\btheta}^2},
\]
the first condition becomes
\[
\sigma_{\btheta}^2
<
\frac{m}{m-1}\sigma_0^2,
\]
and the second condition becomes
\[
|K_{\btheta}-K_{\mathrm{old}}|
<
\sqrt{
\frac{
m\sigma_0^2-(m-1)\sigma_{\btheta}^2
}{
m(m-1)\sigma_s^2
}
}.
\]
Thus the moment-controlled region is
\begin{equation}
\Theta_m
=
\left\{
(K_{\btheta},\sigma_{\btheta}):
\sigma_{\btheta}^2
<
\frac{m}{m-1}\sigma_0^2,
\quad
|K_{\btheta}-K_{\mathrm{old}}|
<
\sqrt{
\frac{
m\sigma_0^2-(m-1)\sigma_{\btheta}^2
}{
m(m-1)\sigma_s^2
}
}
\right\}.
\label{eq:app_theta_m_current}
\end{equation}
For every \(\btheta\in\Theta_m\), we have
\[
\E[|X_t(\btheta)|^m]<\infty.
\]
\paragraph{Verification of the advantage moment condition in LQR.}
The LQR advantage has finite moments of every finite order under the
old policy. Indeed, by \eqref{eq:app_advantage_polynomial_bound},
\[
|A^{\pi_{\bthetaold}}(s,a)|^q
\le
C_q\left(1+|s|^{2q}+|Z|^{2q}\right)
\]
holds for any fixed $q<\infty$. Since $s\sim\mathcal N(0,\sigma_s^2)$ and
$Z\sim\mathcal N(0,1)$, both terms on the right-hand side have finite
expectation. Hence,
\[
\E_{\bthetaold}
\left[
|A^{\pi_{\bthetaold}}(s_t,a_t)|^q
\right]
<\infty
\qquad
\text{for every }q<\infty.
\]
Taking $q=m(m+\eta)/\eta$ verifies the required moment-transfer
condition.

\subsection{Additional details for Figure~\ref{fig:motivation_lqr}}
\label{app:lqr_figure_details}
We use the scalar linear--quadratic instance with dynamics and reward
parameters \(A=0.8\), \(B=1\), \(\gamma=0.95\), \(q=1\), and
\(\lambda=1\), together with noise levels \(\sigma_w=0.3\) and
\(\sigma_0=0.4\), and old feedback gain \(K_{\mathrm{old}}=-0.5\).
These values define a simple, well-conditioned toy instance, rather
than a special physical system.  The choice \(A=0.8\) gives moderately
persistent state dynamics, while \(B=1\) fixes the control scale.
Setting \(q=\lambda=1\) keeps the state and action penalties on the
same scale, making the closed-form value, advantage, and surrogate
expressions easy to interpret.  The old closed loop is stable because
\(A+B K_{\mathrm{old}} = 0.3\), so that the discounted value and the
stationary state law are both well defined. Table~\ref{tab:lqr_variance_edges_multiple_m} summarizes the parameter choices used in Figure~\ref{fig:motivation_lqr}. 

\begin{table}[t]
\centering
\caption{Variance boundary values and corresponding population target values for
fixed \(K_{\btheta}=-0.45\).  Here \(K_{\mathrm{old}}=-0.5\),
\(\sigma_0=0.4\), \(\sigma_s\simeq0.52\), \(D\simeq0.048\), and
\(\Lambda\simeq2.299\).}
\label{tab:lqr_variance_edges_multiple_m}
\begin{tabular}{c c c c c}
\toprule
Moment order \(m\)
& \(k\)
& Boundary for \(\sigma_{\btheta}^2\)
& Boundary for \(\sigma_{\btheta}\)
& \(\mathcal L(\btheta)\) \\
\midrule
\(2\)
& \(0.500\)
& \(0.3186\)
& \(0.5645\)
& \(-0.1816\) \\
\(3\)
& \(0.333\)
& \(0.2380\)
& \(0.4878\)
& \(-0.0889\) \\
\(1/0.7 \approx 1.429\)
& \(0.700\)
& \(0.5315\)
& \(0.7290\)
& \(-0.4262\) \\
\(1/0.9 \approx 1.111\)
& \(0.900\)
& \(1.5989\)
& \(1.2645\)
& \(-1.6530\) \\
\bottomrule
\end{tabular}
\end{table}

\section{Implementation Details}
\label{app:details_for_algorithm}
\subsection{Algorithm}
\paragraph{Overview} Algorithm~\ref{alg:opo} summarizes the implementation of \textsc{OPO}. The procedure follows the standard PPO pipeline: collect rollouts, estimate advantages, split samples into minibatches, and update the policy and value networks. The only modification is the policy loss, where PPO's fixed clipping radius is replaced by the sample-wise radius \(\varepsilon_t\) defined in \eqref{eq:eps_rule_piecewise}. Thus, \textsc{OPO} retains the implementation structure of PPO while introducing order-statistic tail control through the policy objective.
\begin{algorithm}[t]
\caption{Ordered Policy Optimization (OPO)}
\label{alg:opo}
\begin{algorithmic}[1]
\STATE \textbf{Initialize:} policy $\pi_{\btheta}$, value network $V_\phi$, 
hyperparameters $\kappa, \sigma^{\star},k^{\star}$, minibatch size $B$,
value loss and policy entropy coefficients $c_1,c_2$
\WHILE{not converged}
    \STATE Collect
    $\mathcal D=\{(s_t,a_t,\mathcal{R}_t)\}_{t=1}^{N}$ under policy $\pi_{\btheta}$.
    \STATE Set $\pi_{\btheta_{\old}} \leftarrow \pi_{\btheta},\quad
    V_{\phi_{\old}} \leftarrow V_\phi$
    \STATE Estimate 
    $\widehat A(s_t,a_t)$ following \citet{schulman2015high}
    \STATE Set $\widehat{\mathcal{R}}_t \leftarrow
    V_{\phi_{\old}}(s_t)+\widehat A(s_t,a_t)$
    \STATE Split $\mathcal D$ into minibatches 
    $\{\mathcal B\}$ of size $B$
    \FOR{each minibatch $\mathcal B$}
        \STATE
        \[
        \mathcal{L}_{\mathrm{OPO}}
        \leftarrow
        -\frac{1}{|\mathcal B|}
        \sum_{t\in \mathcal B}
        \left\{
        \frac{\pi_{\btheta}(a_t\mid s_t)}
        {\pi_{\btheta_{\old}}(a_t\mid s_t)}
        \cdot \widehat A(s_t,a_t)
        -
        \frac{|\widehat A(s_t,a_t)|}{2\varepsilon_t}
        \cdot
        \left[
        \frac{\pi_{\btheta}(a_t\mid s_t)}
        {\pi_{\btheta_{\old}}(a_t\mid s_t)}
        -1
        \right]^2
        \right\},
        \]
        where $\varepsilon_t$ is defined in \eqref{eq:eps_rule_piecewise}.
        \[
        \mathcal L_e
        \leftarrow
        \frac{1}{|\mathcal B|}\sum_{t\in\mathcal B}
        \mathcal H(\pi_{\btheta}(\cdot\mid s_t)),
        \qquad
        \mathcal L_v
        \leftarrow
        \frac{1}{2|\mathcal B|}\sum_{t\in\mathcal B}
        \left|V_\phi(s_t)-\widehat{\mathcal{R}}_t\right|^2 .
        \]
        \STATE Update loss $\mathcal L \leftarrow
        \mathcal L_{\mathrm{OPO}}+c_1\mathcal L_v-c_2\mathcal L_e$
        \STATE Update parameters
        $\btheta \leftarrow \btheta-\lambda_\theta\nabla_\theta \mathcal L,
        \qquad
        \phi \leftarrow \phi-\lambda_\phi\nabla_\phi \mathcal L$
    \ENDFOR
\ENDWHILE
\STATE \textbf{Output:} 
Policy network $\pi_{\btheta}$
\end{algorithmic}
\end{algorithm}

Intuitively, OPO aims for the tail-controlled regime by assigning likelihood ratio order statistics to a prescribed tail profile. This procedure requires three ingredients: identifying the relevant tail, locating its empirical boundary, and specifying the ordered targets imposed beyond that boundary. Table~\ref{tab:symbols} summarizes the quantities used for these purposes.

\begin{table}[h]
\centering
\caption{Main quantities defining the OPO tail construction.}
\label{tab:symbols}
\begin{tabular}{ll}
\toprule
\textbf{Notation} & \textbf{Role} \\
\midrule
$\kappa$
& Nominal bulk radius and upper bound on the empirical tail boundary \\
$\sigma^\star$
& Scale of the prescribed teacher tail above $u_\eta$ \\
$k^\star$
& Shape parameter governing tail heaviness and the associated moment regime \\
$Q_z$
& Teacher target assigned to the $z$-th retained order statistic \\
\bottomrule
\end{tabular}

\end{table}

\paragraph{Identifying the relevant tail}
The relevant tail depends on the sign of the estimated advantage. Because the surrogate contribution is $r_t\widehat A_t$, increasing $r_t$ is favorable when $\widehat A_t>0$, whereas decreasing $r_t$ is favorable when $\widehat A_t<0$. 

OPO therefore applies the same ordered-tail construction to the upper tail of $r_t$ for positive-advantage samples and to the upper tail of $1/r_t$ for negative-advantage samples. We index these two signed groups by $\eta\in\{+,-\}$ and write their ordered scores as $w^{(\eta)}_{(1)}\leq\cdots\leq w^{(\eta)}_{(B_\eta)}$.

\paragraph{Specifying ordered targets}
Having identified the relevant tail, we next locate the empirical boundary between its bulk and extreme observations. When $M_\eta$ extreme samples are retained from a signed group of size $B_\eta$, the empirical boundary is given by $w^{(\eta)}_{(B_\eta-M_\eta)}$, which can be used to determine the threshold $u_\eta$. In practice, we recommend clamping this value to a reasonable range. For example, one may set \(u_\eta=\max\{1,\min(w^{(\eta)}_{(B_\eta-M_\eta)},1+\kappa)\}\), or use a simpler choice such as clamping to the nominal proximal radius or setting \(u_\eta=1\). In general, we recommend that the resulting threshold lie within \([1,1+\kappa]\).

The order statistic places the boundary immediately below the retained extremes, while $\kappa$ prevents this data-dependent boundary from extending beyond the nominal bulk radius. The resulting $u_\eta$ serves as the point from which the teacher tail is constructed.

Above this boundary, OPO prescribes a generalized Pareto tail with scale $\sigma^\star$ and shape $k^\star$. This choice makes the statistical role of the teacher profile explicit: for a heavy tailed generalized Pareto variable $W$,
\[
\Pr(W>x)\asymp x^{-1/k^\star},
\qquad
\mathbb E[W^p]<\infty
\quad\text{for}\quad
p<1/k^\star.
\]
Hence, choosing
\[
k^\star<\frac{1}{m}
\]
prescribes a tail with finite $m$-th moment. The shape parameter $k^\star$ thus translates the desired moment regime directly into the geometry of the target tail.

The continuous teacher profile must then be converted into targets for the finitely many retained extremes. For rank $z\in\{1,\ldots,M_\eta\}$, OPO uses the plotting probability
\[
p_z=\frac{z-\frac12}{M_\eta}
\]
and assigns the generalized Pareto quantile
\[
Q_z
=
u_\eta
+
\frac{\sigma^\star}{k^\star}
\left[
(1-p_z)^{-k^\star}-1
\right].
\]
Consequently, $u_\eta$ determines where the prescribed tail begins, $(\sigma^\star,k^\star)$ determine its scale and shape, and $z$ determines the target associated with each retained order statistic. For the negative-advantage group, whose ordering is performed on $1/r_t$, the corresponding likelihood ratio target is $Q_z^{-1}$.

\paragraph{Optimizing with assigned targets}
It remains to translate these ordered ratio targets into the policy objective. OPO does so through a sample-specific radius $\epsilon_t$ chosen so that the preferred ratio $r_t^\star$ coincides with the appropriate teacher target. For fixed $\epsilon_t$,
\[
\operatorname*{arg\,max}_{r}
\left[
r\widehat A_t
-
\frac{|\widehat A_t|}{2\epsilon_t}(r-1)^2
\right]
=
1+\epsilon_t\operatorname{sign}(\widehat A_t)
=
r_t^\star.
\]
OPO thereby converts a prescribed tail-moment regime into ordered sample-level ratio targets and, through the local surrogate objective, into the corresponding policy update.

\subsection{Additional diagnostics for the OPO algorithm}
\label{app:opo_diagnostics}

The preceding subsection specifies how OPO maps signed order
statistics to prescribed tail targets and, subsequently, to
sample-specific radii. We now examine whether this construction
behaves as intended in finite samples. In particular, we study whether the prescribed shape parameter $k^\star$
systematically controls the realized likelihood ratio tail, and whether this behavior is robust to nearby choices of the empirical tail threshold.

\paragraph{Control of the realized tail shape.}
We next examine whether the prescribed shape $k^\star$ has its intended
effect on the likelihood ratio distribution. Since $k^\star$ is imposed
on the ordered targets before the policy update, it need not coincide
exactly with a GPD shape parameter fitted to the resulting ratios.
Rather, the relevant question is whether varying $k^\star$
systematically changes the heaviness of the realized tail.

We repeat OPO with $k^\star\in\{0.33,0.49,0.70\}$ and fit the
corresponding GPD shape to the realized relevant tail at nine training
updates. Table~\ref{tab:opo_prescribed_observed_shape} reports the mean
fitted shape for each environment.

\begin{table}[t]
\centering
\caption{Prescribed and observed tail shapes under OPO. Each entry is
the mean GPD shape fitted to the realized relevant likelihood ratio
tail over nine training updates.
\label{tab:opo_prescribed_observed_shape}
}
\begin{tabular}{l c c c}
\toprule
& \multicolumn{3}{c}{Prescribed shape $k^\star$} \\
Environment & $0.33$ & $0.49$ & $0.70$ \\
\midrule
Ant-v4              & 0.243 & 0.387 & 0.559 \\
HalfCheetah-v4      & 0.241 & 0.323 & 0.507 \\
Hopper-v4           & 0.217 & 0.350 & 0.538 \\
Humanoid-v4         & 0.338 & 0.461 & 0.658 \\
HumanoidStandup-v4  & 0.342 & 0.460 & 0.594 \\
Walker2d-v4         & 0.246 & 0.410 & 0.579 \\
\bottomrule
\end{tabular}
\end{table}

The fitted shape increases monotonically with $k^\star$ in every
environment. Thus, OPO does not merely impose a target profile on the
current minibatch: changing the prescribed tail shape produces a
corresponding and systematic change in the realized likelihood ratio
tail. The fitted values need not equal $k^\star$ exactly, but their
ordering is preserved uniformly across all six environments.

\paragraph{Robustness to the empirical tail threshold.}
Finally, we test whether the preceding tail behavior is an artifact of
the particular number of retained extremes. Our default construction
uses
\[
M_\eta
=
\left\lfloor
\min\left\{0.2B_\eta,\,3\sqrt{B_\eta}\right\}
\right\rfloor .
\]
We repeat each tail fit using $20\%$ more exceedances and compare the
resulting nine-point median trajectories of the estimated shape
parameter. The results are reported in
Table~\ref{tab:opo_threshold_sensitivity}.

\begin{table}[t]
\centering
\caption{Sensitivity of the fitted tail-shape trajectory to the
empirical threshold. The nearby construction retains $20\%$ more
exceedances than the default rule.}
\label{tab:opo_threshold_sensitivity}
\begin{tabular}{l c c c c}
\toprule
Tail
& Main $\widehat k$ range
& Exceedances
& Trajectory correlation
& $\max |\Delta \widehat k|$ \\
\midrule
$r_t$, $\widehat A_t \geq 0$
& 0.219--0.519 & 84--125 & 0.975 & 0.067 \\
$1/r_t$, $\widehat A_t < 0$
& 0.340--0.755 & 52--106 & 0.923 & 0.129 \\
\bottomrule
\end{tabular}

\end{table}

The two threshold choices yield strongly aligned shape trajectories:
their correlations are $0.975$ for the positive-advantage tail and
$0.923$ for the negative-advantage tail, with maximum absolute
deviations of $0.067$ and $0.129$, respectively. Hence, the observed
tail-shaping behavior is preserved under a nearby perturbation of the
empirical threshold.

Taken together, these diagnostics support the intended behavior of
OPO's tail construction. Minibatch subdivision reduces extremal
resolution without inflating the induced radii, the prescribed
parameter $k^\star$ systematically controls the realized tail
heaviness, and this effect persists under nearby threshold choices.

\subsection{Further Details}
We fix the hyperparameters to match prior work \citep{xie2024simple}, rather than optimizing them for \textsc{OPO}. All experiments were conducted using NVIDIA RTX 2080 GPUs. The table below summarizes the hyperparameters used.
\begin{table}[t]
\centering
\caption{Detailed hyperparameters used.}
\label{tab:spo-hyperparameters}
\begin{tabular}{lcc}
\toprule
\textbf{Hyperparameters} 
& \textbf{Atari 2600} 
& \textbf{MuJoCo} \\
\midrule
Optimizer                                      & Adam    & Adam    \\
Number of workers                             & 8       & 8       \\
Total steps                                   & 10M     & 10M     \\
Horizon                                       & 128     & 256     \\
Batch size                                    & 1024    & 2048    \\
Number of minibatches                                  & 4       & 4       \\
Number of tasks                                 & 42       & 6       \\
Minibatch size                               & 256     & 512     \\
Update epochs                                 & 4       & 10      \\
Learning rate                                 & 0.00025 & 0.0003  \\
Learning rate decay                           & Linear  & Linear  \\
Discount factor $\gamma$                      & 0.99    & 0.99    \\
GAE parameter $\lambda$                       & 0.95    & 0.95    \\
Value loss coefficient $c_1$                  & 0.5     & 0.5     \\
Entropy loss coefficient $c_2$                & 0.01    & 0.0     \\
Probability ratio hyperparameter $\kappa$   & 0.2     & 0.2     \\
\bottomrule
\end{tabular}
\end{table}

\section{Broader Experimental Results}
\subsection{MuJoCo-v4}
\label{app:mujoco-further-results}
\paragraph{Overall results.}
\begin{figure}[t]
    \centering
    \includegraphics[width=0.95\linewidth,
        trim={0.3cm 0.4cm 0.2cm 0.3cm},
        clip]{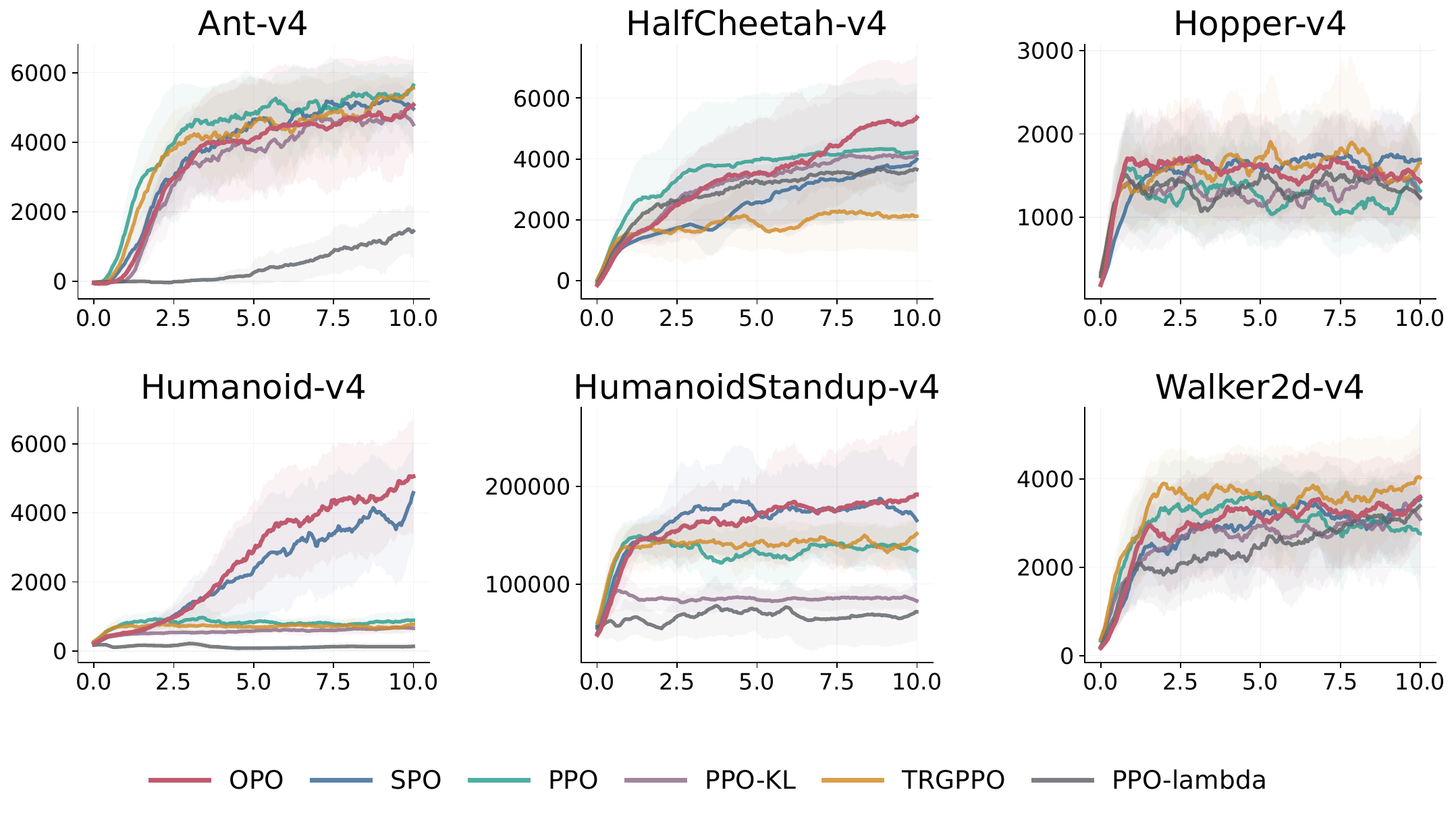}
    \caption{
    MuJoCo-v4 learning curves under the architecture-selected protocol. Solid curves
    denote averages over seeds, and shaded regions indicate
    across-seed variability. }
    \label{fig:mujoco_best_arch}
\end{figure}

Figure~\ref{fig:mujoco_best_arch} reports the corresponding learning
curves. \textsc{OPO} achieves competitive final performance
on most tasks. The results show that replacing fixed pointwise clipping with order-statistic proximal
targets can improve optimization without changing the rollout,
advantage-estimation, or optimizer components of the PPO pipeline.

\paragraph{Computational Cost.}
Further, we also report wall-clock times for the MuJoCo experiments. Table~\ref{tab:mujoco_time} shows that \textsc{OPO} is computationally comparable to PPO and SPO across all environments. Although the order-statistic update introduces additional sorting operations, training time remains close to that of the PPO-type baselines, with all methods requiring roughly two hours on average.

\begin{table}[t]
\centering
\caption{
Representative MuJoCo wall-clock training time in hours. Entries report
mean $\pm$ sample standard deviation over seeds. The last row
averages per-environment means.
}
\label{tab:mujoco_time}
\begin{tabular}{lccc}
\toprule
Environment & OPO & PPO & SPO \\
\midrule
Ant-v4              & $1.91 \pm 0.18$ & $1.80 \pm 0.03$ & $2.12 \pm 0.22$ \\
HalfCheetah-v4      & $2.37 \pm 0.67$ & $1.65 \pm 0.01$ & $2.43 \pm 0.34$ \\
Hopper-v4           & $1.79 \pm 0.10$ & $1.60 \pm 0.02$ & $2.04 \pm 0.29$ \\
Humanoid-v4         & $2.44 \pm 0.36$ & $1.92 \pm 0.07$ & $2.39 \pm 0.21$ \\
HumanoidStandup-v4  & $2.65 \pm 0.10$ & $2.37 \pm 0.07$ & $2.17 \pm 0.05$ \\
Walker2d-v4         & $2.03 \pm 0.12$ & $1.61 \pm 0.05$ & $1.83 \pm 0.04$ \\
\midrule
Across-env. avg.    & $2.20 \pm 0.33$ & $1.83 \pm 0.29$ & $2.16 \pm 0.22$ \\
\bottomrule
\end{tabular}
\end{table}

\paragraph{Ablation study of the shape parameter $k$.}
To understand the effect of the GPD shape parameter, we further report
the MuJoCo scores obtained under different values of $k$. Recall that
larger values of $k$ correspond to heavier prescribed ratio tails. 

Crucially, as shown in Table~\ref{tab:mujoco_k_shape_report_transposed} the lighter-tail choice \(k=0.33\) generally gives competitive performance on five of the six environments, while the heavier-tail choices are less favorable. These results are
consistent with the theoretical motivation of OPO: prescribing overly
heavy ratio tails can weaken the stability of the importance-weighted
surrogate. 
\begin{table*}[t]
\centering
\caption{Ablation over the GPD shape parameter $k$ on MuJoCo. Each entry
reports the mean on the first line and the standard deviation on the
second line. Higher is better.}
\label{tab:mujoco_k_shape_report_transposed}
\begin{tabular}{lcccccc}
\toprule
$k$
& Ant-v4
& HalfCheetah-v4
& Hopper-v4
& Humanoid-v4
& \makecell{Humanoid\\Standup-v4}
& Walker2d-v4 \\
\midrule
$0.33$
& \makecell{4821.96 \\ {\tiny $\pm$ 841.13}}
& \makecell{5239.74 \\ {\tiny $\pm$ 1970.91}}
& \makecell{1567.71 \\ {\tiny $\pm$ 165.72}}
& \makecell{4521.40 \\ {\tiny $\pm$ 741.32}}
& \makecell{187300.73 \\ {\tiny $\pm$ 54819.22}}
& \makecell{3412.71 \\ {\tiny $\pm$ 550.91}} \\

$0.49$
& \makecell{4318.59 \\ {\tiny $\pm$ 937.35}}
& \makecell{3563.51 \\ {\tiny $\pm$ 2144.21}}
& \makecell{1498.09 \\ {\tiny $\pm$ 177.21}}
& \makecell{4248.31 \\ {\tiny $\pm$ 271.17}}
& \makecell{216661.41 \\ {\tiny $\pm$ 44528.67}}
& \makecell{3171.71 \\ {\tiny $\pm$ 773.86}} \\

$0.70$
& \makecell{4678.31 \\ {\tiny $\pm$ 916.41}}
& \makecell{2005.49 \\ {\tiny $\pm$ 67.46}}
& \makecell{1543.86 \\ {\tiny $\pm$ 115.41}}
& \makecell{3717.99 \\ {\tiny $\pm$ 573.71}}
& \makecell{177072.27 \\ {\tiny $\pm$ 26268.78}}
& \makecell{2999.32 \\ {\tiny $\pm$ 718.49}} \\
\bottomrule
\end{tabular}
\end{table*}

\begin{figure}[h]
    \centering
    \includegraphics[width=0.9\linewidth,
        trim={0.3cm 0.4cm 0.2cm 0.3cm},
        clip]{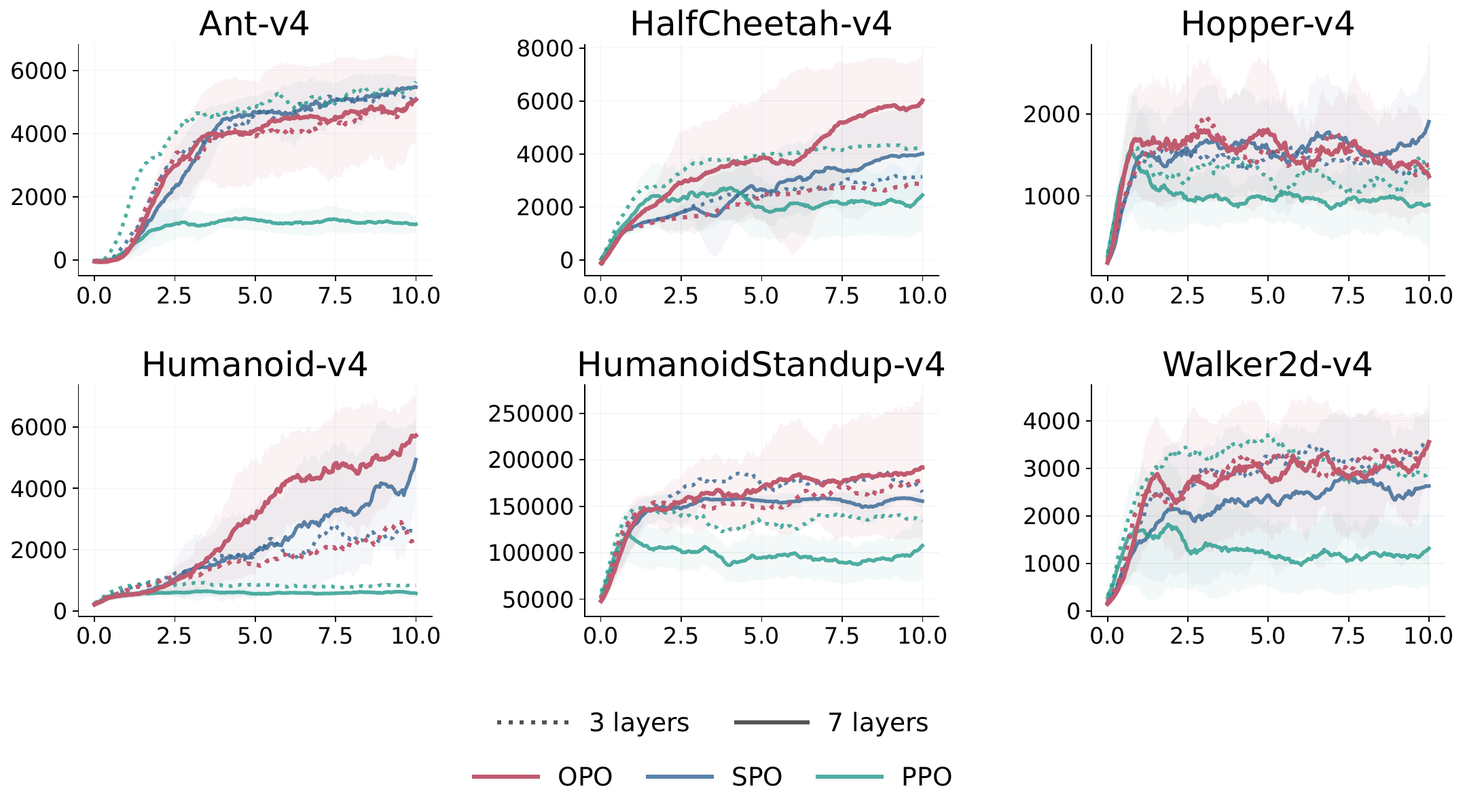}
    \caption{
    MuJoCo-v4 learning curves under different policy-network
    architectures. Solid curves correspond to seven-layer policies,
    while dotted curves correspond to three-layer policies. Colors denote
    different methods. Curves show averages over seeds, and shaded
    regions indicate across-seed variability.
    }
    \label{fig:mujoco_architecture_ablation}
\end{figure}

\paragraph{Ablation study of the policy-network architecture.}
We further examine the sensitivity of each method to the policy-network
architecture. Figure~\ref{fig:mujoco_architecture_ablation} compares
three-layer and seven-layer policies for OPO, SPO, and PPO across the
MuJoCo environments. 

\paragraph{Ablation study of $\kappa$ and $\sigma^{\star}$.}

Table~\ref{tab:mujoco_kappa_scale_report_transposed} reports an ablation over $\kappa$ and the GPD scale $\sigma^\star$. Overall, the results are broadly comparable across settings.
\begin{table*}[h]
\centering
\caption{Ablation over $\kappa$ and GPD scale $\sigma^\star$ on
MuJoCo. Each entry reports the mean on the first line and the standard
deviation on the second line. Higher is better.}
\label{tab:mujoco_kappa_scale_report_transposed}
\begin{tabular}{lcccc}
\toprule
Setting
& Ant-v4
& HalfCheetah-v4
& \makecell{Humanoid\\Standup-v4}
& Walker2d-v4 \\
\midrule
$\kappa=0.2,\ \sigma^\star=0.3$
& \makecell{4902.29 \\ {\tiny $\pm$ 59.21}}
& \makecell{3836.51 \\ {\tiny $\pm$ 1371.68}}
& \makecell{192545.05 \\ {\tiny $\pm$ 21638.89}}
& \makecell{3212.46 \\ {\tiny $\pm$ 196.34}} \\

$\kappa=0.2,\ \sigma^\star=0.1$
& \makecell{4821.96 \\ {\tiny $\pm$ 841.13}}
& \makecell{5239.74 \\ {\tiny $\pm$ 1970.91}}
& \makecell{187300.73 \\ {\tiny $\pm$ 54819.22}}
& \makecell{3412.71 \\ {\tiny $\pm$ 550.91}} \\

$\kappa=0.1,\ \sigma^\star=0.3$
& \makecell{4230.80 \\ {\tiny $\pm$ 443.05}}
& \makecell{4480.91 \\ {\tiny $\pm$ 1935.52}}
& \makecell{177288.03 \\ {\tiny $\pm$ 14523.83}}
& \makecell{3259.16 \\ {\tiny $\pm$ 720.85}} \\

$\kappa=0.1,\ \sigma^\star=0.1$
& \makecell{3618.48 \\ {\tiny $\pm$ 1009.94}}
& \makecell{3586.43 \\ {\tiny $\pm$ 1928.31}}
& \makecell{158855.81 \\ {\tiny $\pm$ 2707.94}}
& \makecell{3040.67 \\ {\tiny $\pm$ 475.03}} \\
\bottomrule
\end{tabular}
\end{table*}

\paragraph{Ablation study of the number of minibatches.}

Table~\ref{tab:mujoco_minibatch_report_transposed} reports an ablation
over the number of minibatches used for each rollout batch. Overall,
performance remains broadly comparable across the tested split counts,
with no single configuration uniformly dominating across environments.

\begin{table*}[h]
\centering
\caption{Ablation over the number of minibatches on MuJoCo. Each entry
reports the mean on the first line and the standard deviation on the
second line. Higher is better.}
\label{tab:mujoco_minibatch_report_transposed}
\resizebox{\textwidth}{!}{
\begin{tabular}{lcccccc}
\toprule
Setting
& Ant-v4
& HalfCheetah-v4
& Hopper-v4
& Humanoid-v4
& \makecell{Humanoid\\Standup-v4}
& Walker2d-v4 \\
\midrule
$2$ minibatches
& \makecell{4479.57 \\ {\tiny $\pm$ 528.75}}
& \makecell{2778.80 \\ {\tiny $\pm$ 955.79}}
& \makecell{1527.27 \\ {\tiny $\pm$ 77.03}}
& \makecell{3465.25 \\ {\tiny $\pm$ 267.99}}
& \makecell{189587.57 \\ {\tiny $\pm$ 43327.78}}
& \makecell{3326.54 \\ {\tiny $\pm$ 261.21}} \\

$4$ minibatches
& \makecell{4821.96 \\ {\tiny $\pm$ 841.13}}
& \makecell{5239.74 \\ {\tiny $\pm$ 1970.91}}
& \makecell{1567.71 \\ {\tiny $\pm$ 165.72}}
& \makecell{4521.40 \\ {\tiny $\pm$ 741.32}}
& \makecell{187300.73 \\ {\tiny $\pm$ 54819.22}}
& \makecell{3412.71 \\ {\tiny $\pm$ 550.91}} \\

$6$ minibatches
& \makecell{4330.91 \\ {\tiny $\pm$ 1155.30}}
& \makecell{5764.36 \\ {\tiny $\pm$ 1056.23}}
& \makecell{1746.36 \\ {\tiny $\pm$ 258.82}}
& \makecell{4589.90 \\ {\tiny $\pm$ 200.05}}
& \makecell{156140.88 \\ {\tiny $\pm$ 3714.68}}
& \makecell{3757.89 \\ {\tiny $\pm$ 215.99}} \\

$8$ minibatches
& \makecell{4568.11 \\ {\tiny $\pm$ 422.11}}
& \makecell{4407.45 \\ {\tiny $\pm$ 1727.63}}
& \makecell{1365.53 \\ {\tiny $\pm$ 119.48}}
& \makecell{4336.45 \\ {\tiny $\pm$ 487.01}}
& \makecell{165167.27 \\ {\tiny $\pm$ 9843.86}}
& \makecell{2695.99 \\ {\tiny $\pm$ 342.23}} \\
\bottomrule
\end{tabular}}
\end{table*}

No split count is uniformly optimal across the six environments:
two minibatches perform best on HumanoidStandup-v4, four on Ant-v4,
and six on the remaining four environments. More generally, the
relative ordering varies across tasks, and the differences are often
comparable to the across-seed variation. These results suggest that
OPO does not rely on a narrowly tuned minibatch count.

\subsection{Atari}
\label{app:atari_further_results}
\paragraph{Quantile summary.}
We report the distribution of relative improvement over PPO across the
Atari environments. Table~\ref{tab:atari_relative_improvement_quantiles}
summarizes the mean and selected quantiles across environments.
\textsc{OPO} yields larger mean and median improvements than SPO, with stronger gains in the upper quantiles.
\begin{table}[t]
\centering
\caption{Relative improvement over PPO on Atari2600. Each row summarizes the
distribution of environment-wise relative improvements across
environments. Higher is better.}
\label{tab:atari_relative_improvement_quantiles}
\begin{tabular}{lcccccccc}
\toprule
Method  & Mean & Min & Q10 & Q25 & Median & Q75 & Q90 & Max \\
\midrule
OPO  & 318.37 & -49.21 & -1.66 & 17.77 & 62.58 & 225.54 & 859.02 & 4150.75 \\
SPO  & 165.45 & -70.47 & -3.29 & 16.41 & 48.95 & 162.89 & 365.28 & 1169.36 \\
\bottomrule
\end{tabular}
\end{table}

\clearpage
\paragraph{Full Learning Curves.}
We present the complete learning curves for the Atari 2600 environments.
\begin{figure}[!h]
    \centering
    \includegraphics[width=0.9\linewidth,
        trim={0.2cm 0.2cm 0.4cm 0.1cm},
        clip]{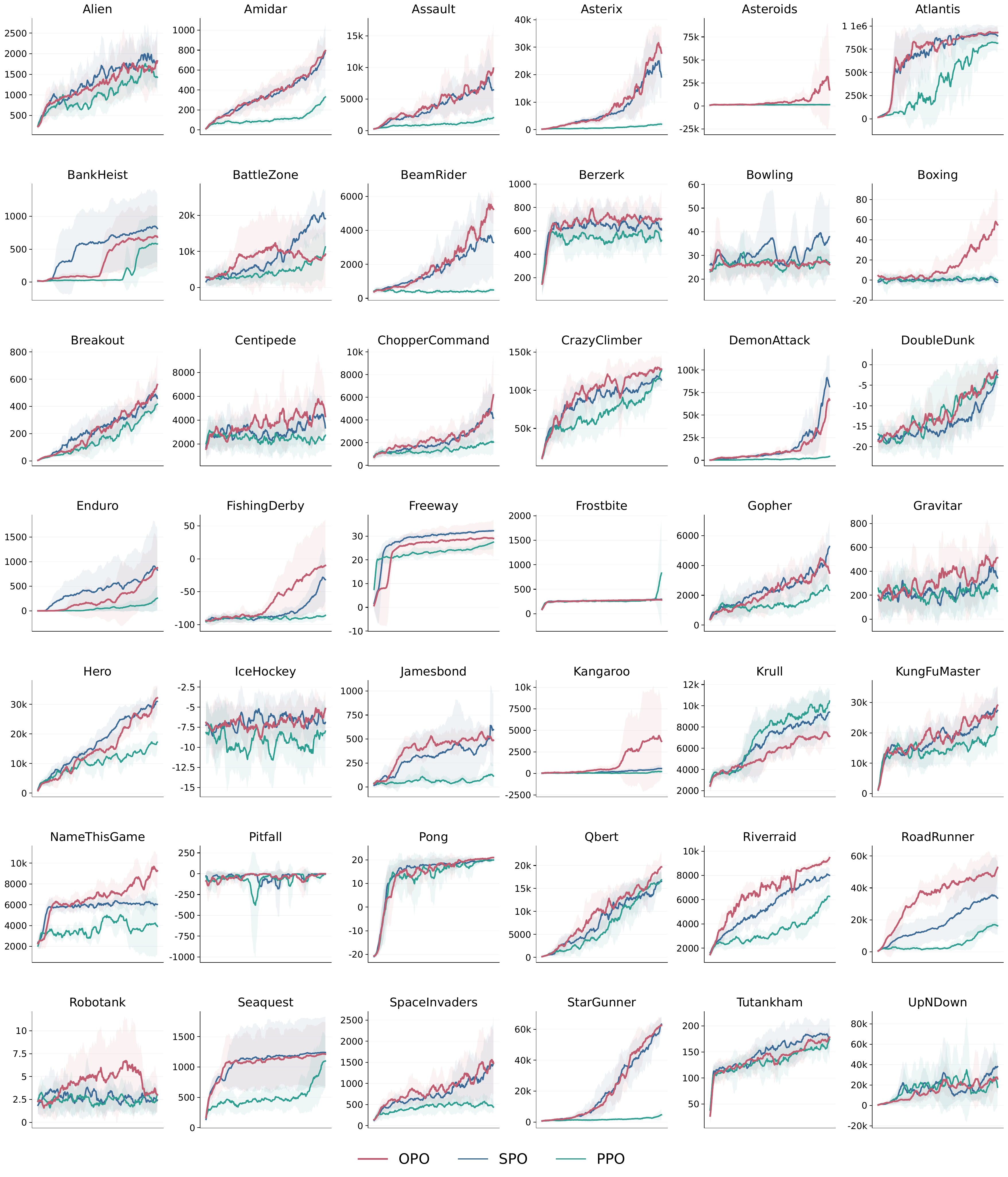}
    \caption{Learning curves on Atari 2600. The mean and standard deviation are shown across 3 random seeds.
    }
    \label{fig:atari_full_learning_curve}
\end{figure}

\clearpage

\newpage
\section*{NeurIPS Paper Checklist}

\begin{enumerate}

\item {\bf Claims}
    \item[] Question: Do the main claims made in the abstract and introduction accurately reflect the paper's contributions and scope?
    \item[] Answer: \answerYes{} %
    \item[] Justification: We explain our contributions and scope in the abstract and introduction.

\item {\bf Limitations}
    \item[] Question: Does the paper discuss the limitations of the work performed by the authors?
    \item[] Answer: \answerYes{} %
    \item[] Justification: We discuss the limitations of the work.

\item {\bf Theory assumptions and proofs}
    \item[] Question: For each theoretical result, does the paper provide the full set of assumptions and a complete (and correct) proof?
    \item[] Answer: \answerYes{} %
    \item[] Justification: We provide full set of assumptions and proof.

    \item {\bf Experimental result reproducibility}
    \item[] Question: Does the paper fully disclose all the information needed to reproduce the main experimental results of the paper to the extent that it affects the main claims and/or conclusions of the paper (regardless of whether the code and data are provided or not)?
    \item[] Answer: \answerYes{} %
    \item[] Justification: We provide experimental details.

\item {\bf Open access to data and code}
    \item[] Question: Does the paper provide open access to the data and code, with sufficient instructions to faithfully reproduce the main experimental results, as described in supplemental material?
    \item[] Answer: \answerYes{} %
    \item[] Justification: The implementation of this work is provided in the following repository: \url{https://github.com/panxulab/Ordered-Policy-Optimization}.

\item {\bf Experimental setting/details}
    \item[] Question: Does the paper specify all the training and test details (e.g., data splits, hyperparameters, how they were chosen, type of optimizer) necessary to understand the results?
    \item[] Answer: \answerYes{} %
    \item[] Justification: We provide all details regarding training and test procedures.

\item {\bf Experiment statistical significance}
    \item[] Question: Does the paper report error bars suitably and correctly defined or other appropriate information about the statistical significance of the experiments?
    \item[] Answer: \answerYes{} %
    \item[] Justification: We compute standard deviations across different seeds.

\item {\bf Experiments compute resources}
    \item[] Question: For each experiment, does the paper provide sufficient information on the computer resources (type of compute workers, memory, time of execution) needed to reproduce the experiments?
    \item[] Answer: \answerYes{} %
    \item[] Justification: We provide detailed computations resources in the Appendix.
    
\item {\bf Code of ethics}
    \item[] Question: Does the research conducted in the paper conform, in every respect, with the NeurIPS Code of Ethics \url{https://neurips.cc/public/EthicsGuidelines}?
    \item[] Answer: \answerYes{} %
    \item[] Justification: We confirm to the code of ethics in every respect.

\item {\bf Broader impacts}
    \item[] Question: Does the paper discuss both potential positive societal impacts and negative societal impacts of the work performed?
    \item[] Answer: \answerNA{} %
    \item[] Justification: There is no societal impact of the work performed.
    
\item {\bf Safeguards}
    \item[] Question: Does the paper describe safeguards that have been put in place for responsible release of data or models that have a high risk for misuse (e.g., pre-trained language models, image generators, or scraped datasets)?
    \item[] Answer: \answerNA{} %
    \item[] Justification: Our work does not pose such risks.

\item {\bf Licenses for existing assets}
    \item[] Question: Are the creators or original owners of assets (e.g., code, data, models), used in the paper, properly credited and are the license and terms of use explicitly mentioned and properly respected?
    \item[] Answer: \answerYes{} %
    \item[] Justification: We provide details of the original owners of assets.

\item {\bf New assets}
    \item[] Question: Are new assets introduced in the paper well documented and is the documentation provided alongside the assets?
    \item[] Answer: \answerYes{} %
    \item[] Justification: We provide our code.

\item {\bf Crowdsourcing and research with human subjects}
    \item[] Question: For crowdsourcing experiments and research with human subjects, does the paper include the full text of instructions given to participants and screenshots, if applicable, as well as details about compensation (if any)? 
    \item[] Answer: \answerNA{} %
    \item[] Justification: The paper does not involve crowdsourcing nor research with human subjects.

\item {\bf Institutional review board (IRB) approvals or equivalent for research with human subjects}
    \item[] Question: Does the paper describe potential risks incurred by study participants, whether such risks were disclosed to the subjects, and whether Institutional Review Board (IRB) approvals (or an equivalent approval/review based on the requirements of your country or institution) were obtained?
    \item[] Answer: \answerNA{} %
    \item[] Justification: Our paper does not involve crowdsourcing nor research with human subjects.

\item {\bf Declaration of LLM usage}
    \item[] Question: Does the paper describe the usage of LLMs if it is an important, original, or non-standard component of the core methods in this research? Note that if the LLM is used only for writing, editing, or formatting purposes and does \emph{not} impact the core methodology, scientific rigor, or originality of the research, declaration is not required.
    \item[] Answer: \answerNA{} %
    \item[] Justification: The core method development does not involve LLMs as any important components.

\end{enumerate}

\end{document}